%% file: main.tex
\documentclass[preprint]{elsarticle}

\usepackage{subcaption}
\usepackage{booktabs}
\usepackage{placeins}
\usepackage{amsmath, amssymb}
\usepackage{ragged2e}
\usepackage{array}
\usepackage{color}
\usepackage{lineno}
\usepackage{makecell}
\usepackage{hyperref}       
\usepackage{url}            
\usepackage{amsfonts}       
\usepackage{nicefrac}       
\usepackage{microtype}      
\usepackage{natbib}

\usepackage{doi}
\usepackage{dcolumn}
\usepackage{bm}
\usepackage{caption}
\usepackage{subcaption}
\usepackage{float}
\usepackage{array}
\newcolumntype{P}[1]{>{\centering\arraybackslash}p{#1}}

\renewcommand{\textcolor}[2]{#2}

\begin{document}

\begin{frontmatter}


\title{\texorpdfstring{History-Aware Neural Operator: Robust Data-Driven Constitutive Modeling of Path-Dependent Materials}{History-Aware Neural Operator}}

\author[UMN]{Binyao Guo}
\author[UMN]{Zihan Lin}

\author[UMN]{QiZhi He\texorpdfstring{\corref{mycorrespondingauthor}}{}}
\ead{qzhe@umn.edu}
\address[UMN]{Department of Civil, Environmental, and Geo- Engineering, University of Minnesota, Minneapolis, MN 55455, USA}

\begin{abstract}
This study presents an  end-to-end learning framework for data-driven modeling of path-dependent inelastic materials using neural operators. The novel framework is built on the premise that the irreversible evolution of material responses, governed by hidden dynamics, can be inferred from observable data.
We develop the History-Aware Neural Operator (HANO), an autoregressive model that predicts path-dependent material responses from 
short segments of recent strain–stress history
without relying on hidden state variables, thereby overcoming the self-consistency issues commonly encountered in recurrent neural network (RNN)-based models. Built on a Fourier-based neural operator backbone, HANO enables discretization-invariant learning. To further enhance its ability to capture both global loading patterns and critical local path dependencies, we embed a hierarchical self-attention mechanism that facilitates multiscale feature extraction. 
Beyond ensuring self-consistency, HANO mitigates sensitivity to initial hidden states, a commonly overlooked issue that can lead to instability in recurrent models when applied to generalized loading paths. By modeling stress–strain evolution as a continuous operator rather than relying on fixed input–output mappings, HANO naturally accommodates varying path discretizations and exhibits robust performance under complex conditions, including irregular sampling, multi-cycle loading, noisy data, and pre-stressed states.  We evaluate HANO on two benchmark problems: elastoplasticity with hardening and progressive anisotropic damage in brittle solids. Results show that HANO consistently outperforms baseline models in predictive accuracy, generalization, and robustness. With its demonstrated capabilities and discretization-invariant design, HANO provides an effective and flexible data-driven surrogate for simulating a broad class of inelastic materials. 

\end{abstract}

\end{frontmatter}

\input{Intro_new}

\input{Preliminaries}

\input{Numerical}

\input{Numerical_damage}
\FloatBarrier

\input{Conclusion}
\section*{Acknowledgment}\label{Ack} 
\textcolor{blue}{The authors gratefully acknowledge the constructive comments and suggestions provided by the anonymous reviewers, which have helped improve the quality and clarity of this manuscript.}
The work is partially supported by QH's Startup Fund and Data Science Initiative (DSI) Seed Grant at the University of Minnesota.

\section*{Data availability}\label{data} 
The training dataset used in this study will be made publicly available at:  \url{https://github.com/IntelligentMechanicsLab/HANO_inelasticity}

\input{appendix}

\newpage

\bibliographystyle{cas-model2-names} 

\bibliography{ref_new,ref}

\end{document}

%% file: Intro_new.tex
\section{Introduction}\label{sec:Introduction}
Constitutive modeling of history-dependent materials presents fundamental challenges in computational solid mechanics. 
\textcolor{blue}{
Traditional phenomenological approaches 
rely on internal variables and evolution equations to describe material behavior
\citep{masi2021thermodynamics,fuhg2024review}. 
Developing such models demands extensive domain expertise for their formulation construction and often involves laborious parameter identification, 
which becomes particularly difficult for materials with complex path-dependent responses such as plasticity, damage, or their combinations.
These challenges have motivated a paradigm shift toward data-driven methods, which aim to learn material behavior directly from experimental or simulation data without imposing strong assumptions about the functional form of the constitutive relationships. 
By replacing intricate model construction with learning from data, data-driven approaches offer greater flexibility in capturing complex behaviors that traditional methods often struggle to represent accurately.
}

Data-driven constitutive modeling methods are commonly categorized into \textit{neural network–based} and \textit{model-free }approaches~\cite{ghaboussi1991knowledge,kirchdoerfer2016data,fuhg2024review,dornheim2024neural}.
Neural network–based methods construct continuous surrogates that map constitutive inputs to target outputs, learning material response behaviors directly from training data. 
Since the pioneering work of Ghaboussi et al.~\cite{ghaboussi1991knowledge}, a wide range of neural network architectures have been explored for path-dependent material modeling, 
such as feedforward neural networks~\cite{huang2020machine,linka2021constitutive}, recurrent neural networks (RNNs) \cite{mozaffar2019deep,gorji2020potential}, 
temporal convolutional networks (TCNs) \cite{wang2022deep,abueidda2021deep},
among others. 
By contrast, model-free approaches bypass any explicit functional form and directly utilize observed data to characterize material behavior~\cite{kirchdoerfer2016data}, 
often functioning as unsupervised manifold learning strategies~\cite{he2020physics_lcdd,he2021deep,fuhg2024review}. 
These model-free methods have been applied to nonlinear elasticity
\cite{nguyen2018data,he2021manifold}, inelasticity~\cite{eggersmann2019model,bartel2023data}, damage mechanics~\cite{carrara2020data, carrara2021data}, etc. 
To mitigate the challenges associated with high-dimensional and noisy data, recent efforts use autoencoders for low-dimensional embeddings~\citet{he2021deep} or employ isometric embeddings to construct denoised constitutive manifolds~\citet{bahmani2023distance}.
However, model-free methods often require extensive data and struggle to generalize beyond observed conditions, which limits their applicability for complex, path-dependent material behaviors unless supported by prior structural knowledge or large datasets~\cite{fuhg2024review}.

\subsection{Recurrent neural networks}
Among neural network–based models, the introduction of Recurrent Neural Networks (RNNs)~\cite{Goodfellow-et-al-2016} and their variants
marked another significant advancement in capturing history-dependent material behavior. RNNs possess internal states that can naturally capture loading history without explicitly defining internal variables or evolution equations. 
\citet{mozaffar2019deep} demonstrated that RNNs can predict path-dependent plasticity without relying on traditional assumptions of yield criteria and flow rules. \citet{gorji2020potential} successfully applied RNNs with gated recurrent units to model anisotropic hardening and multi-axial stress-strain responses in cellular materials. \citet{abueidda2021deep} extended this approach to thermo-viscoplasticity problems, showing both gated recurrent units and temporal convolutional networks can accurately predict these history-dependent responses. \citet{bonatti2022cp} developed RNN-based surrogate models for crystal plasticity that are compatible with explicit finite element frameworks, while \citet{wu2020recurrent} demonstrated the efficiency of RNN-accelerated multi-scale modeling. Beyond bulk material response, \citet{ frankel2020prediction} pioneered a hybrid approach combining convolutional neural networks with recursive components to predict both the average response and stress field evolution in oligocrystals. Recent work has also focused on improving training methodologies, with \citet{fuchs2021dnn2} introducing a deep reinforcement learning framework for network architecture optimization and \citet{stocker2022novel} developing self-adversarial training schemes to enhance prediction robustness. These approaches demonstrate significant advantages: they naturally handle history dependence through their internal architecture, avoid the complexities of defining yield surfaces and flow rules, and offer remarkable computational efficiency once trained.

Despite these advances, two key challenges limit the practical deployment of RNN-based models. First, as highlighted by \citet{bonatti2022importance}, RNN outputs can depend on the discretization of the loading path, violating the principle that mechanical responses should depend only on the path itself. Approaches like the Linearized Minimal State Cell (LMSC) \citep{bonatti2022importance} mitigate discretization dependence by constraining the state update to an exponential form that approximates a differential equation, ensuring consistent predictions across different time step sizes. Second, conventional RNN‑style surrogates and recurrent neural operators inherit a strong dependency on their initial hidden state. As shown by \citet{kidger2020neural}, these models must seed their internal memory at a prescribed baseline—typically the undeformed, zero‑stress configuration—and then step through every load increment to arrive at the desired prediction time. In practice, however, only partial histories may be recorded, or practical considerations may require that simulations begin at an arbitrary, possibly pre‑stressed point along the loading path. Under such circumstances the hidden state is inevitably mis‑specified, and the ensuing distribution shift can lead to large prediction errors.

\subsection{Neural ODE-based approach}
An alternative paradigm is the internal state variable (ISV)–driven neural network approach, grounded in the neural ordinary differential equation (Neural ODE) framework \citep{chen2018neural}, 
which parameterizes the infinitesimal time-evolution operator of a dynamical system and reconstructs trajectories through numerical integration. 
The core idea is to embed thermodynamically motivated ISVs into the Neural ODE formulation, enabling the model to learn the continuous evolution of inelastic stress responses in a manner that is both data-driven and physically interpretable \citep{jones2022neural}. 
Building on this concept, \citet{jones2022neural} propose ISV-NODE, which updates latent ISVs via a flow-rule network and maps them to stress through a companion stress network, while enforcing Clausius–Duhem consistency and frame invariance to deliver discretization-invariant predictions for both viscoelastic and elastoplastic materials.
\citet{liu2023learning} further advanced this direction by recasting stress and a compact set of internal variables into a continuous-time state-evolution framework. 
Their model employs two lightweight feedforward subnetworks to compute instantaneous rates, combined with a simple forward-Euler discretization that enables autoregressive updates preserving the full loading history, while remaining robust to the choice of time increment.
Similarly, the incremental neural controlled differential equation (INCDE) framework \citep{he2024incremental} introduces an alternative evolution strategy for internal variables, improving the stability of stress predictions.
Despite their discretization robustness, these methods remain fundamentally recurrent—historical information is carried only through hidden states propagated from the initial condition, rather than being explicitly conditioned on the observed stress–strain sequence.
Additionally, the reliance on numerical integration for ODE solvers may introduce computational overhead during online prediction.

\subsection{\textcolor{blue}{Thermodynamic Consistency in Data-Driven Constitutive Modeling}}

\textcolor{blue}{Another important class of data-driven constitutive modeling approaches focuses on embedding thermodynamic consistency constraints to ensure physical meaningfulness and improve generalization capabilities. These methods offer the significant advantage of guaranteeing that model predictions adhere to fundamental thermodynamic laws, providing theoretical rigor and enhanced reliability under limited training data conditions. Physics-informed constraints in these frameworks are typically implemented through three main strategies: (1) selecting network architectures that inherently satisfy constraints; (2) incorporating penalty terms in loss functions to prevent constraint violations; and (3) employing posterior rejection strategies that filter optimized models based on predefined acceptance/rejection criteria \cite{fuhg2024review}.}

\textcolor{blue}{Thermodynamics-based Artificial Neural Networks (TANNs) proposed by Masi et al.~\cite{masi2021thermodynamics,masi2022multiscale} represent a prominent methodology in this domain, which directly encodes thermodynamic principles into neural network architectures through dual potential formulations. Under isothermal conditions, this approach combines Helmholtz free energy density $F$ and dissipation rate potential $D$ using two independent neural networks: one for predicting internal variable increments and another for predicting Helmholtz free energy density. This architectural design ensures that stress updates strictly adhere to thermodynamic laws, particularly the second law of thermodynamics. Similarly, several studies~\cite{klein2022polyconvex,thakolkaran2022nn,kalina2022automated, rosenkranz2023comparative, flaschel2025convex} have also employed input convex neural networks to ensure material stability and thermodynamic consistency.}


\textcolor{blue}{However, potential-based approaches face several practical challenges. The determination of appropriate structure and number of internal variables often remains empirical. Additionally, incorporating convexity constraints or dissipation potentials frequently requires complex optimization schemes, automatic differentiation, and numerical integration procedures. Moreover, these methods may exhibit sensitivity to temporal discretization due to reliance on explicit ODE solvers for state evolution~\cite{he2024incremental}, which can compromise the robustness objectives of operator learning approaches.}

\subsection{Proposed methodology, major contributions, and paper organization}

To address both discretization sensitivity and initial-state dependence in modeling path-dependent inelastic materials, we propose a novel \textit{end-to-end data-driven constitutive modeling} framework based on recent advances in neural operator learning~\citep{li2020fourier,kovachki2023neural,lu2021learning}, termed the  history-aware neural operator (HANO).
HANO builds on the Fourier neural operator (FNO) ~\citep{li2020fourier} and adopts an autoregressive formulation inspired by Neural\,ODEs. 
By leveraging fast Fourier transforms, the embedded FNO component can capture global and long-range interactions in the frequency domain,
enabling the model to effectively handle multi-scale phenomena and 
loading histories sampled at varying resolutions. \textcolor{blue}{To best of authors' knowledge, this is the first application of neural operator architectures to history-dependent constitutive modeling.}
Through its autoregressive structure, 
HANO infers future stresses directly from short segments of observable strain–stress history, eliminating the need for hidden state variables or integration from an undeformed reference configuration. 
This design mitigates self-consistency issues and reduces sensitivity to assumed initial conditions.  
The FNO backbone is further enhanced with a U-shaped architecture~\citep{ronneberger2015u} and attention mechanisms~\citep{vaswani2017attention}, allowing the model to capture both global patterns (via spectral representations) and localized path dependencies (via convolutional paths and attention). As a result, the proposed HANO approach delivers accurate, discretization-invariant stress predictions and can be robustly applied to scenarios involving partial or irregularly sampled loading histories.

To the best of the authors’ knowledge, there has been limited investigation into  the sensitivity of data-driven constitutive models to the initial state—particularly in scenarios where prediction must begin from a pre-stressed configuration without access to the full preceding deformation history. 
The authors contend that addressing this challenge is essential for achieving robust generalization and for ensuring that  learned models faithfully capture the underlying irreversible material behavior.
Moreover, in contrast to previous studies that primarily utilize neural operators to map spatially varying material parameters or microstructures to stress responses~\cite{he2024sdeeponet, zhang2024operator}, the present work explores the application of neural operators for history-aware constitutive modeling.
Specifically, it focuses on learning sequential strain–stress operator mappings for path-dependent inelastic materials.
The main contributions of this work are summarized as follows:
\begin{enumerate}
    \item Develop a novel history-aware neural operator for end-to-end data-driven modeling of general path-dependent materials, eliminating the need for explicit internal variable construction.

    \item 
    Demonstrate the enhanced robustness of the proposed method in handling sensitivities to discretization and initial-state conditions.

    \item 
    Demonstrate that the proposed model maintains high predictive accuracy under complex loading conditions, including scenarios with noisy inputs and extended multi-cycle extrapolation.

\end{enumerate}

The remainder of this study is structured as follows.  Section \ref{sec:problem-setup and RNN} reviews the fundamentals of 
data-driven constitutive modeling for path-dependent materials, emphasizing the strengths and limitations of RNN-based approaches.
Section \ref{sec:method} introduces the proposed HANO framework,
detailing its architecture, autoregressive formulation, and training scheme.
In Section \ref{sec:elasto-plastic}, we demonstrate HANO’s ability to accurately model one-dimensional elastoplastic behavior with kinematic hardening, addressing key shortcomings of RNN-based models. 
Section~\ref{sec:damage} extends the evaluation to brittle materials with progressive anisotropic damage, 
analyzing the impact of architectural choices, attention placement, and history window length on predictive performance.
Finally, Section~\ref{sec:conclusion} summarizes the key findings and offers concluding remarks.

%% file: Preliminaries.tex
\section{Data-Driven Constitutive Modeling}\label{sec:problem-setup and RNN}

In this section, we introduce the core concept of data-driven constitutive modeling for path-dependent materials, which entails predicting the subsequent stress state based on historical stress and strain data. We begin by outlining the thermodynamic foundations, which indicate that irreversible material behavior is governed by evolution equations associated with internal state variables (ISVs). We then review the architecture of recurrent neural networks (RNNs), which have been widely used for their ability to model temporal dependencies through hidden states. However, we identify two major limitations of standard RNNs in constitutive modeling: lack of self-consistency and sensitivity to initial conditions. 
These challenges motivate this study to develop a more consistent and robust operator learning framework tailored to the complexities of inelastic material behavior, which is detailed in  Section \ref{sec:method}.

\input{Sec2_1}

\subsection{Data-driven constitutive learning for path-dependent materials}
\label{sec:problem-setup}
This section presents the general formulation of data-driven constitutive modeling for path-dependent materials. This formalism serves to highlight the key challenges that motivate our proposed approach, while also establishing connections to the existing methods, such as RNN, discussed previously.

We begin with a continuous-time description to characterize the history-dependent stress-strain material behavior.   
The continuous mapping of the constitutive relation defined in Eqs. \eqref{eq:gen_constitutive} -  \eqref{eq:gen_evoln}
can be expressed as:
\begin{equation}
\begin{aligned}
\mathcal{G}: \left( \boldsymbol{\varepsilon}(t+\Delta t), \mathbf{z}(t) \right) \mapsto \boldsymbol{\sigma}(t+\Delta t) \\
\quad \text{with} \quad \mathbf{z}(t) = \mathcal{E} \left( \{\boldsymbol{\varepsilon}(\tau), \boldsymbol{\sigma}(\tau)\}_{0 \le \tau \le t} \right)
\end{aligned}
\label{eq:con_mapping_1}
\end{equation}
where
$\boldsymbol{\varepsilon}(t)$ and $\boldsymbol{\sigma}(t)$ denote the strain and stress at current time $t$, $\{\boldsymbol{\varepsilon}(\tau), \boldsymbol{\sigma}(\tau)\}_{0 \le \tau \le t}$ represents the prior strain–stress history,
\textcolor{blue}{$\Delta t$ represents an arbitrary time increment},
and $\mathcal{G}$ represents the evolution operator that maps the historical loading path into a latent state $\mathbf{z}(t)$ encoding the material memory.
The goal is to predict the future stress state $\boldsymbol{\sigma}(t+\Delta t)$ based on
the given future strain state $\boldsymbol{\varepsilon}(t+\Delta t)$
and the current internal states $\mathbf{z}(t)$.

Recognizing that the evolution of internal states depends on the historical strain–stress response, the mapping can be compactly written as:
\begin{equation}
      \mathcal{G}_z:
  \bigl(\{\boldsymbol{\varepsilon}(\tau), \boldsymbol{\sigma}(\tau)\}_{0 \le \tau \le t},\,
    \Delta \boldsymbol{\varepsilon}(t)
        \bigr)
  \;\mapsto\;
  \boldsymbol{\sigma}(t+\Delta t),
  \label{eq:con_mapping_2}
\end{equation}
where $\Delta\boldsymbol{\varepsilon} = \boldsymbol{\varepsilon}(t+\Delta t) - \boldsymbol{\varepsilon}(t)$ is the strain increment, and
the subscript 'z' indicates that the mapping implicitly relies on internal latent states.

In practical experiments or numerical simulations, strain and stress measurements are typically recorded at discrete manner. This leads to the following discrete representation:
\begin{equation}
    \boldsymbol{\varepsilon}_n := \boldsymbol{\varepsilon}(t_n),\quad
  \boldsymbol{\sigma}_n := \boldsymbol{\sigma}(t_n),\quad
  \Delta\boldsymbol{\varepsilon}_{n+1} := \boldsymbol{\varepsilon}_{\,n+1} - \boldsymbol{\varepsilon}_{n}.
  \label{eq:discrete_rep}
\end{equation}
The data-driven mapping is then formulated to 
predict the next stress state $\boldsymbol{\sigma}_{n+1}$ from the strain–stress history sequence up to step $n$ and the strain increment $\Delta\boldsymbol{\varepsilon}_{n+1}$:
\begin{equation}\label{eq:full-history}
  \mathcal{G}_z:\;
  \Bigl(\{\boldsymbol{\varepsilon}_i,\,\boldsymbol{\sigma}_i\}_{i=0}^{n},\;
        \Delta\boldsymbol{\varepsilon}_{n+1}\Bigr)
  \;\mapsto\;
  \boldsymbol{\sigma}_{n+1}.
\end{equation}

Many studies have shown that it is often sufficient to use only a few history steps of strain and stress to accurately capture path-dependent behavior~\cite{mozaffar2019deep,gorji2020potential,he2022thermodynamically}.
Hence, in practice, to reduce memory and computational costs, we typically consider only the most recent $k$ steps:
\begin{equation}
  \mathcal{G}_z:\;
  \Bigl(\{\boldsymbol{\varepsilon}_i,\,\boldsymbol{\sigma}_i\}_{i=n-k+1}^{n},\;
        \Delta\boldsymbol{\varepsilon}_{n+1}\Bigr)
  \;\mapsto\;
  \boldsymbol{\sigma}_{n+1},
\label{eq:map_reduced-history}
\end{equation}
where $k$ is the chosen window size. This approach assumes that the current stress state is primarily influenced by the most recent $k$ strain and stress responses, consistent with the axiom of fading memory in rational thermodynamics \cite{truesdell1966rational,horstemeyer2010historical}.

This mapping in Eq. \eqref{eq:map_reduced-history} provides the mathematical foundation for many existing data-driven constitutive models, which aim to learn the underlying path-dependent material behavior from the recent history of observable states. 
However, how to accurately represent the evolution operator \textcolor{blue}{$\mathcal{G}_z$} in data-driven modeling remains an open challenge, leading to a variety of approaches.

\subsection{Recurrent Neural Networks (RNNs)}\label{sec:RNN}
This section reviews the idea of recurrent neural networks (RNNs), 
one of the most widely used and successful neural architectures for data-driven modeling of path-dependent constitutive behaviors~\cite{mozaffar2019deep,fuhg2024review,he2022thermodynamically,he2023machine}.
This is because the sequential architecture in RNNs naturally offers 
the ability to record and update hidden states across time steps, making them particularly suitable for modeling path-dependent materials characterized by the evolution of ISVs.
Specifically, RNNs approximate the mapping 
Eq. \eqref{eq:map_reduced-history} 
 by introducing an evolving hidden variable $\mathbf{h}$,  which allows the model to capture essential history-dependent features in the dataset.

\begin{figure}[htbp]
    \centering
    \includegraphics[width=0.8\textwidth]{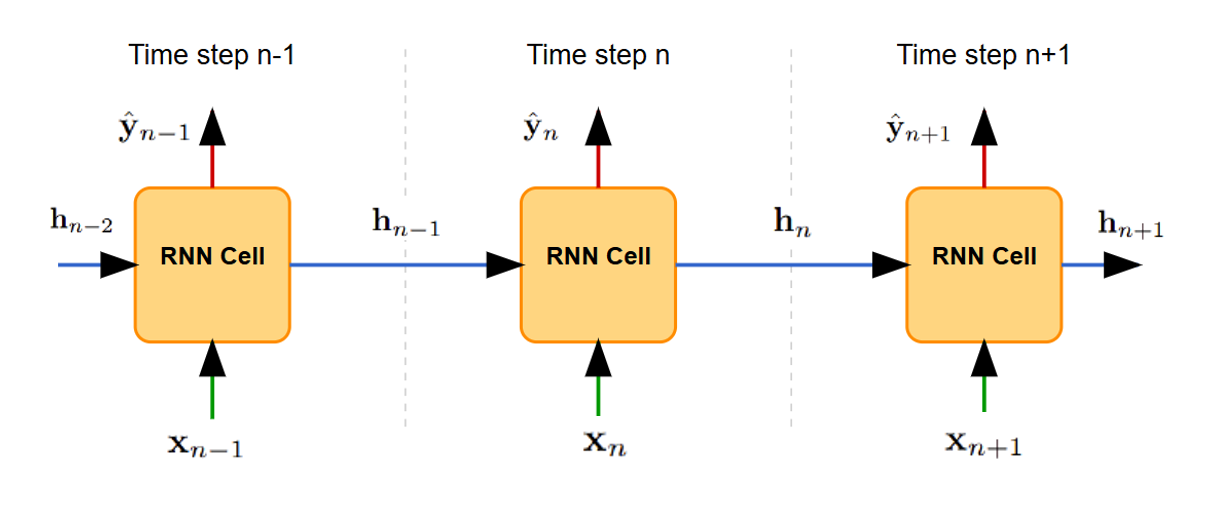}
    \caption{Unfolded RNN architecture across time steps. The blue horizontal arrows represent the hidden state flow between time steps, which enables the network to maintain memory of previous inputs. At each time step, the RNN cell accepts an input vector (green arrows) and the previous hidden state, then produces an output vector (red arrows) and updates the hidden state.}
    \label{fig:rnn_architecture}
\end{figure}

A generic RNN model defines a generic mapping:
\[
   f_{\theta}:\;(\mathbf{x}_n,\mathbf{h}_{\,n-1})\;\longmapsto\;(\hat{\mathbf{y}}_n,\mathbf{h}_{\,n}),
\]
where $\mathbf{x}_n$ represents the input vector at time step $n$, $\mathbf{h}_{\,n-1}$ is the hidden (memory) state before processing step $n$, 
and $\theta$ denote the collection of all trainable parameters (e.g., weights and bias) of the network.
This RNN cell produces an output vector $\hat{\mathbf{y}}_n$ and updates the hidden state to $\mathbf{h}_{\,n}$ after processing.  Figure~\ref{fig:rnn_architecture} illustrates this recurrent structure across multiple time steps.
The hidden state serves as a compact summary of all relevant history up to step \(n\), enabling the RNN to model history‑dependent behavior.

Formally, forward propagation within a simple (tanh‑activated) RNN cell is expressed: For $n = 1,...,N$,
\begin{subequations}\label{eq:rnn_forward}
\begin{align}
\mathbf{h}_{\,n} &= \tanh\!\bigl(\mathbf{W}_{hh}\,\mathbf{h}_{\,n-1}
                        + \mathbf{W}_{xh}\,\mathbf{x}_n
                        + \mathbf{b}_h \bigr), \label{eq:rnn_forward_h}\\[2pt]
\hat{\mathbf{y}}_n &= \mathbf{W}_{hy}\,\mathbf{h}_{\,n} + \mathbf{b}_y,
                     \label{eq:rnn_forward_y}
\end{align}
\end{subequations}
Equations~\eqref{eq:rnn_forward_h} and \eqref{eq:rnn_forward_y} detail how the hidden state \(\mathbf{h}_n\) and the output \(\hat{\mathbf{y}}_n\) are computed. As shown in Figure~\ref{fig:rnn_architecture}, the RNN cell processes information sequentially across time steps, with the hidden state acting as the network's memory mechanism. 
By iterating these updates over \(n\), the hidden state implicitly encodes the historical information needed for subsequent predictions. 
All parameters (\(\mathbf{W}_{hh}\), \(\mathbf{W}_{xh}\), \(\mathbf{W}_{hy}\), \(\mathbf{b}_h\), \(\mathbf{b}_y\)) are shared across every time step. 
These network parameters are trained by comparing the predicted output $\hat{\mathbf{y}}_n$ with the ground truth $\mathbf{y}_n^{*}$.

\subsection{Pitfalls of RNN-based constitutive modeling}\label{sec:pitfalls}
While  RNN-based approaches are well-suited for  learning time-dependent data, they exhibit two notable limitations in the context of path-dependent material constitutive modeling as follows.
\begin{itemize}
\item \textit{Limitation 1: Lack of consistency with respect to resolution (discretization dependence)}

Given the nature of continuum mechanics, classical constitutive relations and  their associated evolution equations are inherently continuous and should be independent of the discretization of the loading path.
However, many recent studies have shown that standard recurrent network architectures often lose this self-consistency property~\cite{bonatti2022importance} due to the finite discretization of the input sequence (e.g., the strain-stress history in Eq. \eqref{eq:discrete_rep}).
In this case, RNNs update the hidden states at every input step, 
so changing the increment size alters the sequence of hidden states—ultimately resulting in different predicted stress trajectories.
This behavior violates the fundamental principle that material response should remain invariant under reparameterization of the loading path.

\item \textit{Limitation 2: Susceptibility to initial condition (hidden state dependence)}

Another key limitation of conventional RNN-based constitutive models is their susceptibility on the initial hidden state.
During \textit{training phase}, the hidden vector $\mathbf{h}_{\,0}$ is typically initialized from a predefined reference configuration, most often in the undeformed zero-stress state, and is subsequently updated  through the recursive relation in Eq. \eqref{eq:rnn_forward}.
Since $\mathbf{h}_{\,n}$ implicitly stores the entire loading history, 
any deviation from this training baseline during the \textit{inference/prediction phase} can lead to significant distribution shift. 
For instance, when RNN predictions are employed in scenarios where the test specimen is already pre‑stressed or when the early portion of the strain-stress history is unavailable,
the model must evolve from an incorrect $\mathbf{h}_{\,0}$. 
In such cases, the RNN is unable to
fast-forward
to the correct internal memory or reconstruct the missing increments, resulting in misalignment with the underlying evolution ODEs (Eq. \eqref{eq:gen_evoln}). 
As discussed in Section \ref{sec:start_data}, the RNN model can result in large prediction errors when the model is forced to begin from mid‑sequence data rather than from an undeformed baseline.
\end{itemize}

The lack of self-consistency (Limitation 1)  in RNNs has been recognized in several recent studies~\cite{bonatti2022importance,he2022thermodynamically,liu2023learning,he2024incremental}, 
and various strategies have been proposed to mitigate this issue. 
For example, the Linearized Minimal State Cell (LMSC) is used to mitigate discretization dependence by constraining the RNN state update using an exponential formulation \citep{bonatti2022importance}. The random walk strategy \citep{he2023machine} improves generalization by introducing variability in step sizes and elastic–plastic transitions during data generation.
Other efforts include appending a hidden-state correction layer to GRU models \citep{he2023machine}, and employing recurrent-type feed-forward networks 
evaluated at each time instance~\citep{liu2023learning}, thereby adopting a continuous-time formulation to alleviate sensitivity to path discretization.

In contrast, Limitation 2 (the sensitivity to initial conditions) has received significantly less attention, despite its critical role in ensuring robust deployment of data-driven models in real-world applications. 
This \textit{generalizability} limitation becomes particularly relevant for in-situ characterization and finite element simulations of pre-loaded solid structures, where the initial conditions may differ from those seen during training.

To address both limitations, we propose a novel data-driven constitutive modeling framework—the history-aware neural operator (HANO)—an autoregressive architecture built upon the Fourier neural operator (FNO) and enhanced with hierarchical attention mechanisms. As elaborated in the next section, 
HANO inherently achieves discretization invariance through spectral learning and, unlike RNN-based approaches, eliminates the need for hidden state variables by directly utilizing short segments of observable strain–stress history as input.


\section{History Aware Neural Operator (HANO) for Material Modeling}\label{sec:method}

In this section, we begin by reviewing the fundamental concepts of neural operators in Section \ref{sec:NO}, including the FNO model and its enhanced U-Net variant, 
with particular attention to their resolution-invariance properties. 
Section \ref{sec:attn} discusses the integration of attention mechanisms into FNO to enable multiscale feature extraction.
Finally, Section \ref{sec:architecture} presents the overall architecture and training procedures of the proposed HANO framework.

A key distinction of HANO compared to RNN-based models lies in its treatment of temporal history. Instead of encoding temporal dependencies through evolving hidden states, 
HANO directly ingests a short window of recent strain–stress observations along with the forthcoming strain increment to predict the next stress state. 
This autoregressive formulation eliminates the need for zero-stress initialization or hidden state reconstruction, allowing the model to make accurate predictions from any point along the loading path using only observable data.

This design offers both flexibility and robustness, enabling accurate predictions even when forecasting starts mid-path or under irregular sampling conditions. 
It forms the foundation of the HANO framework, which provides a practical and generalizable surrogate model for data-driven constitutive modeling of complex inelastic materials.

\subsection{Neural operators}\label{sec:NO}
As introduced in Section \ref{sec:problem-setup}, the goal of data-driven constitutive modeling is to learn a continuous mapping that captures the history-dependent stress–strain behavior of materials, traditionally described by constitutive relations involving latent state variables governed by ODEs. Specifically, the mapping defined in Eq.~\eqref{eq:con_mapping_1} characterizes the material response based on the loading history, where the internal state $\mathbf{z}(t)$ encodes the path dependence.

The key idea behind neural operators is to directly approximate such history-to-stress mappings in an end-to-end fashion. That is, neural operators learn a continuous mapping
\begin{equation}
\mathcal{G}: \mathcal{A} \to \mathcal{V} \to \mathcal{U}
\label{eq:14}
\end{equation}
between infinite-dimensional function spaces $\mathcal{A}$ and $\mathcal{U}$,  
where $\mathcal{V}$ denotes a latent space. We assume that this latent space, learned from training data, can implicitly represent the evolving internal process governed by hidden dynamics $\mathcal{E}$ (as described in Eq. \ref{eq:con_mapping_1}).
This assumption enables the model to avoid explicitly defining or tracking ISVs, i.e., operating in a \textit{hidden variable-free manner}, while still capturing complex history-dependent material behaviors.

For completeness, we briefly introduce the general formulation of neural operators following \cite{kovachki2023neural,li2020fourier}.
Unlike traditional neural networks that focus on finite-dimensional spaces, the deep-learning architectures of neural operators are  designed to learn mappings between infinite-dimensional function spaces. 
Thus, the mapping in Eq. \eqref{eq:14}, $\mathcal{G}:\mathcal{A}\to\mathcal{U}$, can be approximated through a parameterized neural operator $\mathcal{G}_{\phi}$, with trainable parameters drawn from a finite-dimensional space $\phi \in \mathbb{R}^p$. 

Let $\mathcal{D} \subset \mathbb{R}^d$ be a bounded open interval, and $\mathcal{A}:=\mathcal{A}(\mathcal{D}; \mathbb{R}^{d_{a}})$, $\mathcal{V}:=\mathcal{V}(\mathcal{D}; \mathbb{R}^{d_{v}})$ and $\mathcal{U} := \mathcal{U}(\mathcal{D}; \mathbb{R}^{d_{u}})$
be separable Banach spaces of functions taking values in $\mathbb{R}^{d_{a}}$, $\mathbb{R}^{d_{v}}$ and $\mathbb{R}^{d_{u}}$, respectively.
The parametric mapping $\mathcal{G}_{\phi}$ can be formulated as 
\begin{equation} 
\label{eq:NO_formulation}
\mathbf{u} = \mathcal{G}_{\phi}(\mathbf{a};\phi):= \mathcal{Q} \circ \mathcal{L}_{L} \circ \cdots \mathcal{L}_{\ell} \circ \cdots \circ \mathcal{L}_{1} \circ \mathcal{P}(\mathbf{a}), 
\end{equation} 
where $\mathbf{a}(x) \in \mathcal{A}$ and $\mathbf{u}(x) \in \mathcal{U}$ are the input and output functions, respectively, with $x \in \mathcal{D}$ 
representing generalized coordinates defined in the parameter domain \(D\).
In Eq. \eqref{eq:NO_formulation}, $\mathcal{P}$ and $\mathcal{Q}$ are lifting and projection operators, respectively. \textcolor{blue}{While conceptually designed to map between function spaces, they are practically implemented as multilayer perceptrons that operate on finite-dimensional discretized representations}
Also, $\mathcal{L}_{\ell} = \rho\bigl(\mathcal{W}_{\ell}+\mathcal{K}_{\ell}\bigr)$ represents the operator of the $\ell$th hidden layer, i.e., $\mathcal{L}_{\ell}: \mathcal{V} \to \mathcal{V}$. 
In each layer, $\rho$ denotes a point-wise non-linearity activation function,  $\mathcal{W}_{\ell}$ is a local linear operator, and 
$\mathcal{K}_{\ell}$ is an integral kernel operator.

This formulation highlights the three-stage operator architecture—consisting of lifting, iterative integral-kernel transformation, and projection—which enables rich feature extraction and accurate recovery of physical outputs. 
The lifting operator $\mathcal{P}$ elevates the input function $\mathbf{a}(x)$ to a higher-dimensional feature space, i.e., $\mathbf{v}_0(x) = \mathcal{P}(\mathbf{a}(x))$. Subsequently, the hidden representation $\mathbf{v}_0(x)$ undergoes processing through $L$ neural operator layers, $\mathbf{v}_0 \rightarrow \mathbf{v}_1 \rightarrow \cdots \rightarrow \mathbf{v}_{L}$, where $ \mathbf{v}_{\ell} \in \mathcal{V}$. 
Finally, $\mathbf{v}_{L}$ is transformed back to the target output space via the projection operator, $\mathbf{u}(x)=\mathcal{Q} (\mathbf{v}_{L}(x))$.

Analogous to the neural ODE framework~\cite{chen2018neural} and graph network-based simulators~\cite{sanchez2020learning}, the recursive neural operators through layers $\mathcal{L}_{1}$ to $\mathcal{L}_{L}$ can be interpreted as a higher-order discrete approximation of the hidden continuous-time dynamics (characterized by a underlying ODE system), which describe the latent evolution of $\mathbf{v}_t(x)$ from time $t$ to $t+\Delta t$.  In this setting, the operators $\mathcal{P}$ and $\mathcal{Q}$ can thus be viewed as the encoder and decoder, respectively.

There exist multiple architectural choices for designing each layer operator $\mathcal{L}_{\ell}$~\cite{kovachki2023neural}, including graph kernel  network \cite{li2020GNO}, Fourier Neural Operators (FNOs)~\cite{li2020fourier}, and other variants such as DeepOnet~\cite{lu2021learning} and implicit FNO~\cite{you2022learning}, among others.
While any of these architectures can be readily integrated into the proposed HANO framework, in this work, we adopt the FNO architecture due to its computational efficiency, discretization invariance, and strong performance in capturing nonlocal dependencies, as demonstrated in \cite{kovachki2023neural}.

\subsubsection{Fourier neural operator}\label{sec:fourier-layer}
In the Fourier Neural Operator (FNO)~\cite{li2020fourier}, the generic hidden layer
$\mathcal{L}_{\ell}$ in~\eqref{eq:NO_formulation} is replaced by the
Fourier layer $\mathcal{L}^{\text{F}}_{\ell}$, while the overall
operator skeleton remains unchanged.  
The layer $\mathcal{L}^{\text{F}}_{\ell}$ performs global convolution in the
frequency domain of $\mathcal{D}$, thereby efficiently capturing non-local
correlations.  
For an incoming feature representation
$\mathbf{v}_{\ell-1}(x)$, the layer  applies a
Fast Fourier Transform (FFT) $\mathcal{F}$, multiplies the resulting spectrum by a
learnable complex-valued filter $R_{\ell}$, and finally transforms back via the
inverse FFT $\mathcal{F}^{-1}$.  This procedure defines the kernel-integral
operator
\begin{equation}\label{eq:fno-kernel}
\bigl(\mathcal{K}_{\ell}\mathbf{v}_{\ell-1}\bigr)(x)
  = \mathcal{F}^{-1}\!\Bigl[\,R_{\ell}\,\cdot\,
      \mathcal{F}[\mathbf{v}_{\ell-1}]\,\Bigr](x),
\end{equation}
\textcolor{blue}{The FFT operation in Eq.~\eqref{eq:fno-kernel} transforms the temporal sequence from the time domain to the frequency domain, where each frequency mode captures different temporal scales of the loading history. The learnable complex-valued filter $R_{\ell}$ acts as a frequency-selective operator, enabling the model to emphasize specific temporal patterns based on their relevance to the constitutive response. 
Importantly, the frequency-domain representation makes the learned operator independent of the temporal discretization used during training. Since the learned parameters index physical wavenumbers rather than discrete grid points, the same operator applies consistently across different sampling resolutions, enabling discretization-invariant predictions \cite{li2020fourier}.}  

The
layer output is then obtained by combining this global convolution with a
point-wise linear map and a non-linearity. In summary, the operation of the $\ell$th hidden layer is,
\begin{equation}\label{eq:fno-layer}
\mathbf{v}_{\ell}(x) = \mathcal{L}^{F} (\mathbf{v}_{\ell-1}(x)) 
  = \rho\!\Bigl(\mathcal{W}_{\ell}\,\mathbf{v}_{\ell-1}(x)
    + \bigl(\mathcal{K}_{\ell}\mathbf{v}_{\ell-1}\bigr)(x)\Bigr),
\end{equation}
where $\mathcal{W}_{\ell}$ is a linear operator and $\rho(\cdot)$ denotes the
activation function.  

Each hidden state $\mathbf{v}_{\ell}\in\mathcal{V}$ carries $d_v$ latent channels at every point $x\in D$.  
For instance, on a temporal grid of $k$ steps and $d_a$
physical components, $\mathbf{v}_{\ell}$ is stored as a tensor 
\[
  \mathbf{V}_\ell\in\mathbb{R}^{k\times d_a\times d_v},
\]
whose axes index time, state variable, and latent channel, respectively.

A naive time-domain convolution along the first axis costs
\(
  \mathcal{O}(k^{2})
\)
operations for each of the \((d_a \times d_v)\) channel pairs.
By instead applying a 1-D FFT along that axis,
the per-channel cost drops to \(\mathcal{O}(k\log k)\), so the layer’s total cost
scales as
\(
  \mathcal{O}\!\bigl(d_a d_v\, k\log k\bigr)
\).
Throughout this spectral operation the tensor shape
\(k \times d_a \times d_v\) is preserved.
Consequently,
FNO layers capture multi-scale patterns and long-range interactions within a
single hop, substantially improving the representation of phenomena governed by
intrinsically non-local constitutive relations.
Detailed descriptions of FNO can be found in \cite{li2020fourier, kovachki2023neural}.

\emph{Remark 3.1}: We note that the continuous mapping constructed by the FNO possesses a unique property: it remains independent of any discretization in $\mathcal{D}$~\cite{kovachki2023neural}. This property arises from the Fourier layer, which learns a filter in frequency space whose parameters index the physical wavenumbers rather than grid points, ensuring that the same operator applies 
consistently  
across both coarse or fine discretizations. 
In this study, we leverage this property to learn a resolution-invariant mapping from strain-stress history windows to future stress state, 
which is to be further discussed in Section \ref{sec:architecture}.

\subsubsection{U-Net enhanced FNO}\label{sec:u-fno}

While the Fourier layer excels at capturing global trends and low-frequency information, it can suffer from reduced accuracy in representing fine–scale features due to the inherent regularization of the truncated Fourier basis. 
To address this limitation, \citet{wen2022u} introduced the \emph{U-Fourier} layer, which augments a Fourier layer with a parallel U‐Net style convolutional path. 
This combination enhances the network’s ability to capture local, high‐frequency information that a purely Fourier‐based path might overlook.

Given an intermediate hidden feature $\mathbf{v}_{\ell-1}$, the U-Fourier layer produces the output: 
\begin{equation}\label{eq:ufno-layer}
\mathbf{v}_{\ell}(x)= \mathcal{L}^{UF}(\mathbf{v}_{\ell-1}(x)) = \rho\!\Bigl(
      \mathcal{K}(\mathbf{v}_{\ell-1})(x)\;+\;
      \mathcal{U}(\mathbf{v}_{\ell-1})(x)\;+\;
      \mathcal{W}(\mathbf{v}_{\ell-1})(x)
\Bigr),
\end{equation}
where $\rho$ is a point-wise nonlinearity applied to the summed element-wise outputs from three branches: (1) Fourier branch: $\mathcal{K}(\mathbf{v}_{\ell-1})$  captures global and low-frequency features in the frequency domain via a learnable spectral convolution, as described in Eq. \eqref{eq:fno-layer}; (2) U-Net branch: $\mathcal{U}(\mathbf{v}_{\ell-1})$ is an encoder–decoder network with skip connections that extracts multiscale, high-frequency details; (3) Linear branch: $\mathcal{W}(\mathbf{v}_{\ell-1})$ applies linear operator that preserves original feature content and facilitates cross-channel interactions.

In addition, \citet{wen2022u} demonstrated that the architecture composed of a sequence of standard Fourier layers ($\mathcal{L}^F$) followed by U‐Fourier layers ($\mathcal{L}^{UF}$) yields  improved performance. 
This improvement likely stems from the early Fourier layers effectively capturing global patterns, while the subsequent U‐Fourier layers enhance the representation of local features and high‐frequency components. 
Consequently, this sequential architecture is adopted in the construction of the proposed HANO model (Section \ref{sec:architecture}), and our numerical investigations also confirm this observation.


\input{Sec_3_Attn}

\subsection{History-aware neural operator and its implementation}
\label{sec:architecture}

\begin{figure}[htbp]
  \centering
  \includegraphics[width=0.7\textwidth]{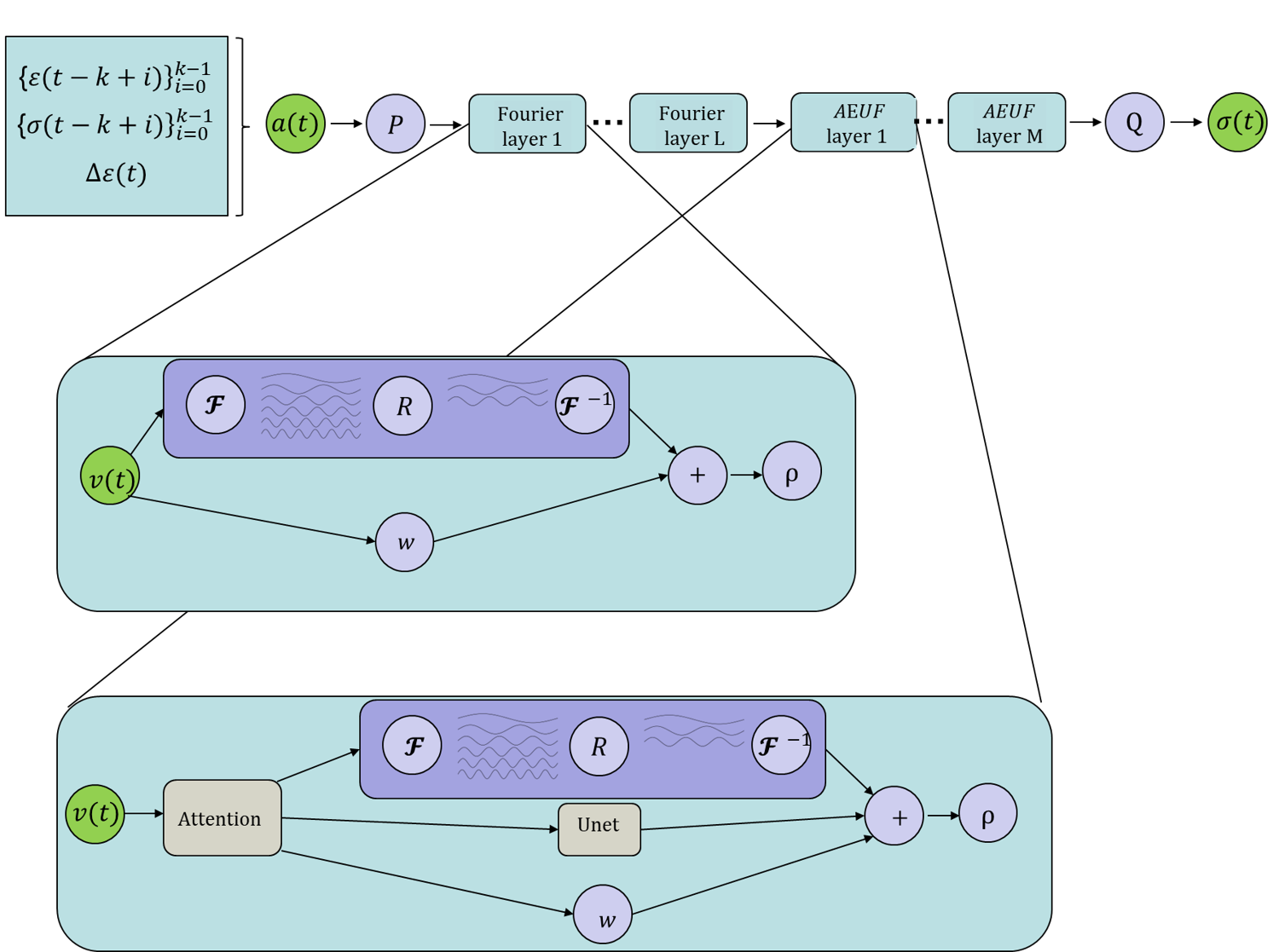}
  \caption{Computational graph of the HANO used for stress prediction.}
  \label{fig:structure_model}
\end{figure}

The proposed HANO is an attention-enhanced, multi-stage neural operator architecture designed for data-driven constitutive modeling of path-dependent materials.
Formally, HANO approximates the continuous mapping for the continuous operator mapping for the constitutive relation $\mathcal{G}_z$ defined in Eq. \eqref{eq:con_mapping_2},
expressed as:
\begin{equation}
  \mathcal{G}_{\phi}^{H} (\mathbf{a})=
  \mathcal{Q}  \circ  \!\Bigl(
    \underbrace{\mathcal{L}^{AEUF}_M \circ \cdots \mathcal{L}^{AEUF}_m\cdots\circ \mathcal{L}^{AEUF}_1 }_{\text{Attention–U‑Fourier layers}}
    \,\circ\,
    \underbrace{\mathcal{L}^{F}_{L} \circ \cdots \mathcal{L}^{F}_{\ell}\cdots \circ \mathcal{L}^{F}_{1}}_{\text{Fourier layers}}
    \,  \Bigr) \circ\, \mathcal{P}\bigl(\mathbf{a}\bigr),
  \label{eq:overall-architecture}
\end{equation}
where $\mathbf{a}(x)$ denotes the input function for $x \in \mathcal{D}$, the lifting and projection operators
$\mathcal{P}$ and $\mathcal{Q}$ are as described in Eq. \eqref{eq:NO_formulation},
the layers $\mathcal{L}^{\text{F}}_{\ell} (l=1,...,L)$ correspond to the standard Fourier layers defined in Eq. \eqref{eq:fno-layer},
and $\mathcal{L}^{\text{\textit{AEUF}}}_{m} (m=1,...,M)$ denote the AEUF layers.
The overall architecture of the HANO model is illustrated in \autoref{fig:structure_model}.

As shown in Eq. \eqref{eq:overall-architecture}, 
HANO extends the standard neural operator framework by interleaving multiple standard Fourier layers with AEUF layers.
Specifically, it leverages 
$\mathcal{L}^{\text{AEUF}}$ layers to further refine the latent representations produced by the preceding Fourier layers, through  coupling the global spectral context with attention-guided, multiscale detail recovery, as discussed in Section \ref{sec:attn}.
The embedded self-attention module dynamically re-weights the strain–stress history, allowing the model to selectively focus on the most relevant segments for next-step prediction.
Within each $\mathcal{L}^{\text{AEUF}}$ block, a U-Net pathway recovers high-frequency and localized features that may otherwise be attenuated by the truncated Fourier basis, while a parallel Fourier branch retains coarse-scale global trends. 
This combination of U-Net, Fourier, and linear branches enables the AEUF layer to effectively fuse low- and high-frequency information in a single pass.
By stacking multiple AEUF layers, 
HANO progressively enhances its capacity to model both long-range dependencies and localized transitions, leading to substantial improvements in path-dependent stress prediction, as demonstrated in Section \ref{sec:comparison_baseline}.
\textcolor{blue}{This HANO architecture serves as the default framework throughout this work and corresponds to the HANO$_3$ variant discussed in Section \ref{sec:attention_placement}.}

The following subsections describe the input–output data structure, training strategy, and implementation details of the HANO framework.

\subsubsection{Data structure of HANO for constitutive modeling}\label{SEC:hano_data} 

Applying HANO to the data-driven constitutive modeling of path-dependent stress-strain prediction described in Eq. \eqref{eq:map_reduced-history}, 
we obtain a discrete representation of the input function $\mathbf{a}$ as 
\begin{equation}    
    \mathbf{A} = \Bigl[\{\boldsymbol{\varepsilon}_i,\,\boldsymbol{\sigma}_i\}_{i=n-k+1}^{n},\;\Delta\boldsymbol{\varepsilon}_{n+1} \Bigr],
    \label{eq:HANO_input}
\end{equation}
and 
the output $\mathbf{u}(x) = \mathcal{G}^{H}_{\phi} (\mathbf{a}(x))$ is the predicted stress $\boldsymbol{\sigma}_{n+1}$. 
Therefore,  $\mathcal{D}$ represents the temporal index domain, corresponding to the sampling points  of the strain–stress history, which we parameterize—albeit with a slight abuse of notation—by the discrete time index \(t_n\).

In the numerical implement of HANO, we rearrange the input $\mathbf{A}$ into a tensor with shape 
\((d_x, d_a, d_c)\),
where $d_x $ represents the sampling points in the temporal domain $\mathcal{D}$ (i.e., $d_x = k$, corresponding to the defined input history length of the strain–stress sequence), $d_a$ denotes the size of the input strain/stress tensor at each time step,
and $d_c$ indicates the number of input channels (e.g., strain and stress). 
This  follows the input setting described in \citep{li2020fourier,kovachki2023neural},
ensuring that every input field appears with the same sequence length, i.e., \(k\), 
which facilitates efficient computation of $\mathcal{P}$.
Note that the strain increment $\Delta\boldsymbol{\varepsilon}_{n+1}$ can be input as a separate channel, replicated $k$ times to match the same sequence length $k$ as the strain–stress history. 

\textcolor{blue}{ When the available stress-strain history is shorter than the required input length \textit{k} (e.g., when loading begins from the virgin undeformed state), we employ zero-padding strategies to maintain consistent input dimensions. Specifically, missing historical data points are padded with zero stress-strain pairs, allowing the model to process loading histories of varying lengths while preserving the temporal structure of the input sequence.}

\emph{Remark 3.2 (Resolution invariance)}: The proposed history-aware neural operator (HANO) inherently provides discretization-invariant predictive capabilities, owing to the 
spectral decomposition in the initial Fourier layers. 
Unlike conventional RNN-based methods that evolve hidden states sequentially at fixed discrete steps, HANO constructs 
operator mappings over continuous function spaces. 
This allows the model to learn path-dependent material behavior in a manner that is \textit{resolution-independent} of time step size or sampling resolution.
As a result, the learned model remains consistent and accurate even under irregular or coarse input discretizations (see Section~\ref{dis_invariance}).

\emph{Remark 3.3 (Hidden variable independence)}: Different from many existing data-driven approaches that rely on reconstructing (physical or nominal) hidden variables encode loading history information \cite{gorji2020potential,liu2023learning, he2024incremental},
HANO directly leverages available strain–stress histories as input. 
This end-to-end formulation allows HANO to operate as a \emph{hidden-state-free} model 
while capable of capturing the underlying dynamics of internal variable (ISV) evolution without explicitly modeling them. \textcolor{blue}{A key practical advantage of this autoregressive formulation is that it relies on observable quantities rather than latent, non-physical hidden states. If a material is under an unknown pre-stressed state, one can apply $k$ small controlled loading increments and record the corresponding stress--strain responses to initialize the model. This procedure can be carried out through controlled experiments on small specimens or, when the data-driven constitutive model is embedded within a finite element solver, through a few trial load steps at the integration point level.} In this way, HANO achieves consistent and robust predictions from any arbitrary point along the loading path. 
This design naturally supports autoregressive forecasting, 
as further demonstrated in \ref{sec:start_data}, where the model accurately predicts path-dependent material behavior even when initialized mid-trajectory.

\input{Sec_3_Training}

%% file: Sec2_1.tex
\subsection{Thermodynamics-based constitutive models}\label{sec:2-1_thermomechanics}
As originally postulated by the internal state variable (ISV) theory \cite{coleman1967thermodynamics},  classical constitutive models for inelastic solids and geomaterials, particularly those involving path-dependent plasticity and damage, can be formulated within the framework of irreversible thermodynamics with ISVs \cite{silhavy2013mechanics,simo2006computational,jones2022neural}.
These models typically result in a system of nonlinear, constrained ordinary differential equations (ODEs) that govern the irreversible evolution of ISVs, which in turn define the constitutive relations linking macroscopic stress and strain and characterize the material behavior under mechanical loading.

For strain-rate independent materials under isothermal adiabatic conditions, the Helmholtz free energy $\Psi$ can be expressed as:
\begin{equation}
    \Psi = \Psi(\boldsymbol{\varepsilon}, \mathbf{z})
\label{eq:cons_phi}
\end{equation}
where $\boldsymbol{\varepsilon}$ is the strain tensor, and $\mathbf{z} = (z_1, ..., z_m)$ represents a collection of $m$ ISVs (or history variables), such as plastic strain, hardening variables, damage variable, etc. 
The Cauchy stress tensor $\boldsymbol{\sigma}$ and the conjugate thermodynamic forces $\mathbf{Y}$ are then obtained as:
\begin{equation}
\boldsymbol{\sigma} = \frac{\partial \Psi (\boldsymbol{\varepsilon}, \mathbf{z})}{\partial \boldsymbol{\varepsilon}}, \mathbf{Y} = -\frac{\partial \Psi (\boldsymbol{\varepsilon}, \mathbf{z})}{\partial \mathbf{z}},
\label{eq:gen_constitutive}
\end{equation}
where $\mathbf{Y} = (Y_1, ..., Y_m)$ denotes the set of thermodynamic forces associated with the internal variable. 

To ensure thermodynamic admissibility, the total dissipation rate must be nonnegative:
\[
\mathcal{D} = \boldsymbol{\sigma} : \dot{\boldsymbol{\varepsilon}} - \frac{d}{dt} \Psi(\boldsymbol{\varepsilon}, \mathbf{z}) \geq 0.
\]
This inequality is also well known as the Clausius-Duhem inequality.
Substituting \( \boldsymbol{\sigma} = \frac{\partial \Psi}{\partial \boldsymbol{\varepsilon}} \) and applying the chain rule for $\frac{d}{dt} \Psi(\boldsymbol{\varepsilon}, \mathbf{z})$, the dissipation rate simplifies to
\begin{equation}
\mathcal{D} = \mathbf{Y} \cdot \dot{\mathbf{z}} \geq 0. 
\label{eq:dissipation}
\end{equation}

To complete the classical constitutive relationship, the evolution laws for the ISVs must be defined to represent the irreversible material processes.
A general rate-type ODE, justified by Truesdell’s equipresence axiom~\cite{truesdell1966rational},
is commonly adopted:
\begin{equation}
\dot{\mathbf{z}} = \mathbf{f}(\mathbf{Y}, \boldsymbol{\sigma}, \mathbf{z}),
\label{eq:gen_evoln}
\end{equation}
such that the dissipation inequality Eq. \eqref{eq:dissipation} holds true at any time $t$.

This provides a natural guideline for formulating evolution equations for ISVs (e.g., associative flow rules, damage growth) that ensure thermodynamic consistency.
From this brief exposition, it is observed that the generic constitutive relation 
$\boldsymbol{\sigma} = \boldsymbol{\sigma} (\boldsymbol{\varepsilon}, \mathbf{z})$ 
for path-dependent materials, as shown in
Eq. \eqref{eq:gen_constitutive}, is
coupled with a system of ODEs, i.e., Eqs. \eqref{eq:dissipation} and \eqref{eq:gen_evoln},
which govern the irreversible evolution of material response. 

It is important to note that the selection of internal state variables (ISVs) and their associated evolution equations is inherently ad hoc, typically constructed to reproduce observed material behavior based on experimental observations and theoretical assumptions~\cite{masi2023evolution}.

\subsubsection*{Example: Elastoplasticity}
Let us consider the theory of elastoplasticity as a representative example. The internal variables include $\mathbf{z} =[\boldsymbol{\varepsilon}^p, \boldsymbol{q}]^T$,
where $\boldsymbol{\varepsilon}^p$ is the plasticity strain tensor and $\boldsymbol{q} $ is a vector of hardening variables. Their evolution laws are expressed as the following rate-independent equations \cite{simo2006computational}:
\begin{equation}
\dot{\boldsymbol{\varepsilon}}^p = \dot{\gamma} \, \mathbf{n}, \quad \dot{\boldsymbol{q}} = \dot{\gamma} \, \mathbf{h}(\boldsymbol{\sigma}, \mathbf{z}),
\end{equation}
where $\mathbf{n} = \frac{\partial \mathcal{F} (\boldsymbol{\sigma}, \mathbf{z})}{\partial \boldsymbol{\sigma}}$ represents the associative plastic flow direction, with $\mathcal{F}$ being the yield function,
$ \mathbf{h}$ defines the hardening law, and 
\( \dot{\gamma} \geq 0 \) is the plastic multiplier ensuring satisfaction of the Karush–Kuhn–Tucker (KKT) condition:
\begin{equation}
    \mathcal{F} (\boldsymbol{\sigma}, \mathbf{z}) \leq 0, \ \dot{\gamma} \geq 0, \ \dot{\gamma} \mathcal{F} = 0.
\end{equation}
The internal variable $\boldsymbol{q}$ typically includes an expansion tracer $\boldsymbol{\alpha}$ (associated with \textit{isotropic hardening}) and a relocation tracer $\boldsymbol{\beta}$ (associated with \textit{ kinematic hardening}), i.e., $\boldsymbol{q} = [\boldsymbol{\alpha},\boldsymbol{\beta}]^T$. 

Collectively, these relations define a nonlinear, constrained ODE system:
\begin{equation}
\frac{d}{dt}{\mathbf{z}} = 
\frac{d}{dt}
\begin{bmatrix}
\boldsymbol{\varepsilon}^p \\
\boldsymbol{q}
\end{bmatrix}
= \dot{\gamma} \begin{bmatrix}
\mathbf{n}(\boldsymbol{\sigma}, \mathbf{z}) \\
\mathbf{h}(\boldsymbol{\sigma}, \mathbf{z})
\end{bmatrix},
\end{equation}
subject to $\mathcal{F} \leq 0, \ \dot{\gamma} \geq 0, \ \dot{\gamma} \mathcal{F} = 0$. 
More details of the elastoplasticity model and its numerical implementation are provided in Section \ref{sec:elasto-plastic} and \ref{appx:A}. 

\bigskip
We remark that
identifying and calibrating classical constitutive laws for general path-dependent inelastic materials remains challenging due to 
the inherent nonlinearity and the implicit dependence of evolution laws on internal state variables (ISVs). Moreover, the evolution equations governing ISVs,
such as
Eq. \eqref{eq:gen_evoln}, require \textit{a priori} assumptions about the functional forms of material behavior.
These assumptions can introduce modeling bias and limit flexibility. 
ISVs are often high-dimensional, not directly observable, and strongly model-dependent, which further complicates interpretation and impedes generalization.
To address these challenges, data-driven constitutive modeling has emerged as a promising alternative \cite{kirchdoerfer2016data,he2020physics_lcdd,lefik2003artificial,mozaffar2019deep,masi2023evolution,he2022thermodynamically,liu2023learning,fuhg2024review,jones2022neural}, as mentioned in Introduction.

%% file: Sec_3_Attn.tex
\subsection{Attention mechanism}\label{sec:attn}

Attention mechanisms have become fundamental components in deep learning architectures \citep{vaswani2017attention}, enabling models to selectively focus on relevant parts of an input sequence. Unlike strictly sequential architectures, attention provides direct access to any position in the sequence, facilitating the capture of both local and long-range dependencies \citep{li2019enhancing}. Its ability to selectively weight different segments of the sequence based on their relevance to the prediction task makes attention especially valuable for complex data with recurring but irregularly spaced patterns. In general, self-attention models employ a query--key--value (QKV) framework. 

Assume  a generic input matrix $X = [x_1, x_2, \dots, x_N] \in \mathbb{R}^{D_x \times N}$, 
where \(D_x\) denotes the feature dimension of each input vector and \(N\) is the sequence length, the self-attention mechanism linearly projects each input vector $x_i$ into three separate vector spaces to produce the query, key, and value representations. We denote the resulting query, key, and value matrices as $Q$, $K$, and $V$, respectively. This linear mapping can be expressed as
\begin{align*}
    Q &= W_q\,X \in \mathbb{R}^{D_k \times N},\\
    K &= W_k\,X \in \mathbb{R}^{D_k \times N},\\
    V &= W_v\,X \in \mathbb{R}^{D_v \times N}\,,
\end{align*}
where $W_q \in \mathbb{R}^{D_k \times D_x}$, $W_k \in \mathbb{R}^{D_k \times D_x}$, and $W_v \in \mathbb{R}^{D_v \times D_x}$ are the learned weight matrices for these linear transformations. Here  
\(D_k\) denotes the dimensionality of the query and key vectors,  
\(D_v\) denotes the dimensionality of the value vectors. We write $Q = [\,q_1, q_2, \dots, q_N\,]$, $K = [\,k_1, k_2, \dots, k_N\,]$, and $V = [\,v_1, v_2, \dots, v_N\,]$, where $q_n$, $k_n$, and $v_n$ denote the query, key, and value vectors corresponding to the input $x_n$, respectively.

For each query vector $q_n$, the self-attention mechanism computes an output vector $a_n$ as a weighted sum of the value vectors:
\begin{equation*}
    a_n \;=\; \sum_{j=1}^{N} a_{nj}\, v_j\,,
\end{equation*}
where the scalar weight $a_{nj}$ indicates the attention that query $q_n$ pays to the $j$th input. These attention weights are typically obtained by applying a softmax function to a similarity score between $q_n$ and each key $k_j$. For example, using a scaled dot-product score $s(k_j, q_n) = (k_j^\mathsf{T} q_n) / \sqrt{D_k}$, the weight becomes $a_{nj} = \operatorname{softmax}\!\big(s(k_j, q_n)\big)$. Substituting this definition into the expression above gives
\begin{equation*}
    a_n \;=\; \sum_{j=1}^{N} \operatorname{softmax}\!\big(s(k_j, q_n)\big)\, v_j\,,
\end{equation*}
which is the expanded form of the attention operation. In matrix form, the collection of all output vectors can be written as 
\begin{equation*}
    H \;=\; V \operatorname{softmax}\!\Big(\frac{K^\mathsf{T}\,Q}{\sqrt{D_k}}\Big)\,,
\end{equation*}
where $H = [\,a_1, a_2, \dots, a_N\,]$ is the matrix of output vectors for all query positions. The $1/\sqrt{D_k}$ factor in the softmax argument is included to mitigate the effect of large dot-product values and improve numerical stability \citep{vaswani2017attention}. 

In essence, the self-attention mechanism uses the above computations to produce \textit{context-dependent} representations that highlight the most pertinent parts of the input. Each output vector $a_n$ is a weighted aggregation of the value vectors in $V$, where the weights emphasize those inputs most relevant to the query $q_n$. This formulation allows the model to learn complex dependencies by dynamically weighting the importance of different positions in the sequence for each prediction. The resulting weighted representation emphasizes the most predictive patterns and naturally adapts to varying structures in the data. 

Prior studies have incorporated attention mechanism into FNO models primarily to improve spatial modeling \citep{peng2022attention, peng2023linear}. 
The demonstrated effectiveness of attention motivates its further application to capture long-range dependencies in history-dependent systems. 
Building on this intuition, we propose introducing the attention component into the U-FNO layer,
enabling it not only to preserve the resolution-independent global feature representation strength but also to adaptively enhance local, high-frequency details through data-driven reweighting. 
This composite building block, termed the \emph{Attention-Enhanced U-Fourier (AEUF) layer}, is presented in Section~\ref{sec:AEUF_layer}.


\subsubsection{Attention–Enhanced U–Fourier (AEUF) layer}
\label{sec:AEUF_layer}

Building on the U–Fourier layer introduced in Section~\ref{sec:u-fno}, we insert a multi-head self-attention operator~$Attn(\cdot)$ before the three-branch processing of each U–Fourier layer (The investigation for the placement of attention is discussed in detail in Section~\ref{sec:attention_placement}).  For a given hidden representation $\mathbf{v}_{m-1}\in\mathcal{V}$, the attention module produces a content-adaptive representation $\tilde{\mathbf{v}}_{m-1}=Attn(\mathbf{v}_{m-1})$ that emphasises the most informative temporal–channel patterns.  The subsequent Fourier, U-Net, and linear branches remain unchanged but now act on~$\tilde{\mathbf{v}}_{m-1}$.  The layer update therefore reads
\begin{equation}\label{eq:aeuf_update}
\begin{aligned}
    \mathbf{y}_{m}(x) & = \mathcal{L}^{AEUF}(\mathbf{v}_{m-1}(x)) = \mathcal{L}^{UF} \circ Attn(\mathbf{v}_{m-1}(x)) \\ 
& = \rho\!\Bigl(      \mathcal{K}\bigl(\tilde{\mathbf{v}}_{m-1}\bigr)(x)+\mathcal{U}\bigl(\tilde{\mathbf{v}}_{m-1}\bigr)(x)+    \mathcal{W}\bigl(\tilde{\mathbf{v}}_{m-1}\bigr)(x)
\Bigr), \quad \forall\,x \in \mathcal{D}
\end{aligned}
\end{equation}


%% file: Sec_3_Training.tex
\subsubsection{Model prediction \& loss function}
\label{sec:autoregressive}

The proposed HANO operates in an autoregressive manner: at every prediction (\textit{inference}) step, it takes a fixed‑length window of the most recent strain–stress pairs together with the forthcoming strain increment and outputs the next stress value, namely
\begin{equation}
\hat{\boldsymbol{\sigma}}_{n+1} = \mathcal{G}^{H}_{\phi}\!\Bigl( \{\!\left(\boldsymbol{\varepsilon}_{i},\boldsymbol{\sigma}_{i}\right)\!\}_{i=n-k+1}^{n}, \;\Delta\boldsymbol{\varepsilon}_{n+1} \Bigr), \label{eq:reduced-history}
\end{equation}
where $n$ denotes the current step, 
and \(k\) denotes the chosen window length. For multi‑step forecasting, we iterate \eqref{eq:reduced-history} recursively: after predicting \(\hat{\boldsymbol{\sigma}}_{n+1}\), we update the strain \(\boldsymbol{\varepsilon}_{n+1}=\boldsymbol{\varepsilon}_{n}+\Delta\boldsymbol{\varepsilon}_{n+1}\), slide the window forward by discarding the oldest pair \((\boldsymbol{\varepsilon}_{n-k+1},\boldsymbol{\sigma}_{n-k+1})\) and appending the newly formed pair \((\boldsymbol{\varepsilon}_{n+1},\hat{\boldsymbol{\sigma}}_{n+1})\), and then supply the next increment \(\Delta\boldsymbol{\varepsilon}_{n+2}\). In short, we set $n \rightarrow n+1$ in Eq. \eqref{eq:reduced-history} in the next step. Repeating this procedure until the desired prediction length yields the complete stress trajectory. 
\textcolor{blue}{We note that, for forecasting in test cases, only the data within the initial window are required.
}

For \textit{model training}, we employ the Mean Squared Error (MSE) loss function to quantify the discrepancy between the predicted next stress value and the ground truth:
\begin{equation}
\mathcal{L}_{\text{MSE}}(\phi)
=\frac{1}{N_{tr}}
\sum_{j=1}^{N_{tr}}
\frac{1}{T_j}
\sum_{i=1}^{T_j}
\bigl(
      \boldsymbol{\sigma}_{i+1}^{(j)}
      -\hat{\boldsymbol{\sigma}}_{i+1}^{(j)}(\phi)
\bigr)^{2},
\end{equation}
where $N_{tr}$ is the number of strain-stress sequences in the training set, $T_j$ is the number of strain–stress data points in the $j$th training sequence,
$\boldsymbol{\sigma}_{i+1}^{(j)}$ denotes the target stress data at the $(n+1)$ step associated with the $j$th sequence, and $\hat{\boldsymbol{\sigma}}_{i+1}^{(j)}(\phi)$ is the corresponding model prediction obtained from~\eqref{eq:reduced-history}.

\textcolor{blue}{Following standard machine learning practice~\cite{goodfellow2016deep}, all strain and stress components are normalized to zero mean and unit variance using the mean and standard deviation computed from the training dataset. The same normalization parameters are consistently applied to both training and test sets, with model predictions denormalized before evaluation.}

\subsubsection{Training scheme}\label{sec:training_schedule}
We employ the Adam optimizer with an initial learning rate of 1e-3 and weight decay of 1e-4 to prevent overfitting. The learning rate follows a step-based decay schedule with a gamma value of 0.5 every 100 epochs, gradually reducing the step size as training progresses to enable fine-grained optimization near the optimal solution. To prevent overfitting and determine the optimal training duration, we implement an early stopping mechanism that monitors the MSE on the validation set. The model's performance is evaluated after each epoch, and training terminates if no improvement is observed for a specified number of consecutive epochs (patience parameter set to 200).

To enhance the model's ability to handle autoregressive prediction, we implement a progressive training strategy with two key components: scheduled sampling \cite{bengio2015scheduled} and noise injection. This strategy addresses the mismatch between training (where ground truth is available) and inference (where the model relies on its own predictions).

\textit{Scheduled sampling}, widely used in training RNNs, addresses the exposure bias problem in autoregressive models by gradually transitioning from teacher forcing to self-reliant prediction. 
At training epoch $e$, 
In this approach, a teacher forcing ratio $\alpha_e \in [0,1]$ will be set.
With probability $\alpha_e$, the network receives the true stress from data as input for the next step, and with probability $1-\alpha_e$, it uses its own predicted stress. In this study, we adopt a scheduled sampling strategy with a monotonically decreasing teacher forcing ratio, linearly reducing from 1.0 to 0.0 in 500 epochs. This progressive reduction allows the model to initially learn stable mappings without the burden of error propagation, while gradually becoming robust to its own prediction errors as training advances.

For \textit{noise injection}, we inject \emph{epoch–dependent} Gaussian perturbations into the stress input during training.
For a clean stress value $\sigma$, the corrupted sample at epoch~$e$ is
\begin{equation}
  \tilde{\sigma}
  \;=\;
  \sigma + \beta,
  \qquad
  \beta \sim \mathcal{N}\!\bigl(0,\;\varsigma_{\text{noise}}^{2}(e)\bigr).
\end{equation}
Inspired by curriculum-learning principles~\citep{morerio2017curriculum}—which advocate \emph{progressively increasing the training difficulty} to enhance model robustness—the noise standard deviation \(\varsigma_{\text{noise}}(e)\) is linearly ramped from \(0.001\) at the start of training
to \(0.020\) by epoch~200, using uniform increments every 50 epochs.
For epochs beyond 200 we keep \(\varsigma_{\text{noise}}\) fixed at \(0.020\).

This training approach serves dual practical purposes. 
In early epochs, we keep inputs nearly clean, allowing the network to learn the fundamental stress-strain relationship without interference. As training progresses, we increase noise levels while simultaneously implementing scheduled sampling to phase out teacher forcing. 
By coupling scheduled sampling with noise injection, the training scheme gradually exposes the network to similar imperfect inputs it will create during inference. As teacher forcing is phased out and controlled noises are added, the model learns to attenuate its own prediction drift while remaining resilient to measurement noise. This joint strategy yields materially more reliable long-horizon stress forecasts, because the network is conditioned to operate under realistic inference conditions well. \textcolor{blue}{Representative training and validation loss curves demonstrating the effectiveness of this training strategy are provided in ~\ref{appx: appendix_loss_curve}.}

\subsubsection{Evaluation metrics}

To evaluate the predictive performance of the neural network models, we employ the Normalized Root Mean Square Error (NRMSE) defined as follows:
\begin{equation}
\operatorname{NRMSE}
=\frac{1}{N_{\mathrm{ev}}}\sum_{j=1}^{N_{\mathrm{ev}}}
\sqrt{%
      \frac{%
        \displaystyle
        \frac{1}{T_j}\sum_{i=1}^{T_j}
        \bigl(
          \boldsymbol{\sigma}_{i}^{(j)}-
          \hat{\boldsymbol{\sigma}}_{i}^{(j)}
        \bigr)^{2}
      }{%
        \displaystyle
        \frac{1}{T_j}\sum_{i=1}^{T_j}
        \bigl(
          \boldsymbol{\sigma}_{i}^{(j)}
        \bigr)^{2}
      }}.
\label{eq:nrmse}
\end{equation}
where $N_{\mathrm{ev}}$ is the number of evaluation sequences, $T_j$ the length associated with the sequence $j$, and $\boldsymbol{\sigma}_{i}^{(j)}$ and  $\hat{\boldsymbol{\sigma}}_{i}^{(j)}$ denote the reference and predicted stresses, respectively.

%% file: Numerical.tex
\section{Numerical Study: 1D Elastoplastic Materials}\label{sec:elasto-plastic}

In this section, we demonstrate the superior accuracy and flexibility of the proposed HANO model  in comparison to  RNN-based models.
Both models are evaluated on a one-dimensional elastoplastic material example with kinematic hardening, where  data preparation and problem setup are described in Section \ref{sec:data_elastoplast}.
We first show in 
Sections  \ref{sec:start_data} and \ref{dis_invariance}
that HANO overcomes the two key limitations of conventional RNN-based approaches
, as discussed in Section \ref{sec:pitfalls}:
sensitivity to discretization and dependence on initial hidden states.
We then  examine the model’s generalization ability and robustness to noise in Sections~\ref{sec:multi_cycle_extrapolation} and~\ref{sec:noise_robustness}, respectively.

In Section \ref{sec:damage}, we extend the evaluation of HANO to a multidimensional anisotropic damage model and perform ablation studies to analyze the architectural components of the neural operator framework.

\subsection{Data preparation \& problem setting}\label{sec:data_elastoplast}
The training and testing datasets used in this verification example are derived from the one-dimensional elastoplastic model with kinematic hardening, 
characterized by the Young's modulus \(E = 200\,\text{GPa}\), the yield stress \(\sigma_{y} = 0.2\,\text{GPa}\), the hardening modulus \(H = 20\,\text{GPa}\).
Details of the constitutive model and its implementation are provided in \ref{appx:A}.
Starting from a stress‐free state (referred to as the \textit{undeformed configuration}), we generate 1,000 distinct loading histories, 
each comprising two complete loading–unloading cycles and  uniformly sampled with 100 increments (i.e., $T_j \equiv 100, \forall j$).
The loading and unloading strain amplitudes \(e_{\text{load}}\) and \(e_{\text{unload}}\) are randomly sampled from the interval \([0.008,\,0.015]\) and \([0.003,\,0.007]\), respectively. 
Unless otherwise stated, all stress‐related quantities in this section are expressed in gigapascals (GPa). 

Out of these 1,000 histories, \(800\) are randomly selected for the training set, and the remaining \(200\) are used for testing. 
For evaluation, these 200 testing cases are divided into two groups  to construct the following two test sets:
\begin{itemize}
    \item Testset I (Full loading history with undeformed start): 
    This set contains 100 cases initialized at zero stress, consistent with the initial conditions used in the training data. Each case includes the complete loading history starting from the undeformed configuration.
    \item Testset II (Partial loading history with deformed start):
    This set contains the remaining 100 cases but they begin from a deformed (nonzero stress) state. To imitate missing or unavailable initial data, we randomly truncate the first 30–50\% of strain increments in each loading history.
\end{itemize}

The HANO used in this study consists of 6 operator layers: $L=3$ standard Fourier layers followed by $M=3$  AEUF layers. 
Each layer employs a spectral resolution of 
$5$ 
Fourier modes and a feature channel width of $w = 64$. The input to this network consists of a sliding window of the most recent $k = 10$ strain–stress pairs, along with the next strain increment. Note that the choice of window size $k$ plays a crucial role in balancing memory length and model complexity. 
A detailed study on the effect of $k$ is presented later in Section \ref{sec:input_length}.

\textcolor{blue}{The effectiveness of the proposed HANO model was evaluated through a few benchmark tests against two variants of RNN-based models:
\begin{itemize}
\item RNN\(_1\): The original RNN model from \citet{gorji2020potential}, designated as RNN\(_1\), which exclusively uses $k$ strain history data as input; 
\item RNN\(_2\): A modified RNN model, designated as RNN\(_2\), whose input layer is revised to receive the same concatenated strain–stress history window and forthcoming strain increment as the HANO model. 
\end{itemize}
}

\textcolor{blue}{This systematic comparison enables us to isolate two key effects: (1) RNN\(_1\) vs RNN\(_2\) comparison isolates the impact of input representation and prediction strategy, where RNN\(_2\) uses identical input structure (strain-stress history + increment) as HANO and conducts prediction autoregressively, and (2) RNN\(_2\) vs HANO comparison isolates the architectural differences between finite-dimensional recurrent networks and operator learning frameworks under identical training protocols and input configurations.}

All data-driven models (HANO and RNNs) are trained with the identical protocol described in Section~\ref{sec:training_schedule}.
\textcolor{blue}{
Table~\ref{tab:dataset_1d} summarizes the dataset configurations used for all numerical studies in this section based on the 1D elastoplastic material, while varying the training and testing configurations to evaluate different aspects of model performance.}

\begin{table}[htb]
\centering
\footnotesize
\renewcommand{\arraystretch}{1.8}
\begin{tabular}{|>{\RaggedRight\arraybackslash}p{2.0cm}|>{\RaggedRight\arraybackslash}p{3.2cm}|>{\RaggedRight\arraybackslash}p{3.8cm}|>{\RaggedRight\arraybackslash}p{3.8cm}|}
\hline
\textbf{Figure \#} & \textbf{Description} & \textbf{Training Dataset} & \textbf{Test Dataset} \\
\hline
Figure \ref{fig:incomplete_history_prediction}  (Section \ref{sec:ref_1d}) 
& Evaluation of initial state sensitivity (comparing RNN$_{1}$,  RNN$_{2}$, and HANO) 
& 
800 histories with kinematic hardening, 100 increments per cycle, zero-stress initialization (Default training dataset described in Section 4.1)
& Single case from Testset II with $\sim$45\% of initial loading history removed, prediction from pre-stressed state ($\sigma \neq 0$, $\varepsilon^p \neq 0$) \\
\hline
Figure \ref{fig:discretization_comparison} (Section \ref{sec:resolution_testset}) 
& Evaluation of resolution sensitivity in test dataset (comparing RNN$_{2}$ and HANO)
& Same configuration as above 
& Different loading paths from Testset I evaluated at multiple discretizations: 60, 70, 80, 90, 100, 120, 130, 140, 150 increments per cycle \\
\hline
Figure \ref{fig:resolution_predictions} (Section \ref{sec:resolution_testset}) 
& Evaluation of discretization sensitivity: Prediction of stress–strain trajectories in representative test cases at different resolutions (comparing RNN$_{2}$ and HANO)  
& Same configuration as above & Single representative test case evaluated at eight different resolutions (60-150 steps) \\
\hline
Figure \ref{fig:variable_resolution_examples} (Section \ref{sec:resolution_training}) 
& 
Evaluation of discretization invariance: 
HANO trained on variable-resolution data and tested across different resolutions
& Variable-resolution training: 1000 cases with randomly distributed resolutions (50-150 steps per cycle) 
& Four test cases with unseen temporal resolutions: 57, 83, 112, and 135 increments per cycle \\
\hline
Figure \ref{fig:five_cycle_example} (Section \ref{sec:multi_cycle_extrapolation}) & Evaluation of extrapolation capability beyond training scope & Same as Figure \ref{fig:incomplete_history_prediction} (Default training dataset described in Section 4.1) & Extended 5-cycle loading sequences for extrapolation assessment \\
\hline
Figure \ref{fig:HANO_noise_comparison} (Section \ref{sec:noise_robustness}) & Evaluation of robustness to noisy input data & Same configuration as above & Gaussian noise added to the test inputs \\
\hline
\end{tabular}
\normalsize
\caption{\textcolor{blue}{
Summary of training and test dataset configurations used in each experiment presented in Section 4, based on the 1D elastoplastic material model introduced in Section 4.1. 
}}
\label{tab:dataset_1d}
\end{table}

\subsection{Verification test: Susceptibility to the initial hidden state}
\label{sec:start_data}
In this test, we  reveal the limitations of  RNNs related to hidden state dependence (Limitation 2 in Section \ref{sec:pitfalls}). 
Specifically, their prediction performance is highly sensitive to  initial states and requires consistency with the training data
In contrast, our HANO model ingests both strain and stress from any accessible segment of the loading history, allowing it to identify the current state and accurately predict future stresses without needing to initialize predictions from the initial undeformed configuration by default.

\subsubsection{Results and discussion}\label{sec:ref_1d}
In this test, all models were evaluated on both Testset I (Full loading history with undeformed start) and Testset II (Partial loading history with deformed start) as described above.

\begin{table}[ht!]
\centering
\begin{tabular}{lccc}
\hline
\textbf{Loading Condition} & \textbf{RNN\(_1\)} \cite{gorji2020potential} & \textbf{RNN\(_2\)} & \textbf{HANO} \\ \hline
Testset I & 0.030 & 0.023 & 0.006 \\
Testset II & 0.359 & 0.106 & 0.004 \\ \hline
\end{tabular}
\caption{Comparison of Normalized Root Mean Square Error (NRMSE) between RNN variants and HANO for  different test sets.}
\label{tab:compare_rnn_hano}
\end{table}

As shown in Table \ref{tab:compare_rnn_hano}, all models achieve satisfactory prediction accuracy on Testset I, where each loading path begins from an undeformed state and contains the complete history.
The NRMSE is
$3\%$ for RNN\(_1\), $2.3\%$ for RNN\(_2\), and $0.6\%$ for HANO. 
HANO substantially reduces the error compared to both RNN variants, highlighting its superior performance in capturing loading histories.

For Testset II, where an initial portion of the loading path is missing, RNN\(_1\) exhibits a dramatic increase in prediction error  to NRMSE=$35.9\%$, 
reflecting its significant reliance on early strain increments for proper hidden state initialization. 
RNN\(_2\), which incorporates both strain–stress history as inputs, attains improved accuracy than RNN\(_1\); however, it still experiences a substantial error increase—fivefold to $10.6\%$.
In contrast, HANO maintains exceptionally high accuracy (NRMSE of $0.4\%$), nearly two orders of magnitude lower than the error of RNN\(_1\), 
 demonstrating the superior performance of neural operator architectures over RNN-based models.
In addition, the comparison between RNN\(_1\) and RNN\(_2\) highlights that using strain–stress history, rather than a single strain sequence, significantly enhances model predictivity.

\begin{figure}[htbp]
    \centering
    \begin{subfigure}[b]{0.32\textwidth}
        \centering
        \includegraphics[width=\textwidth]{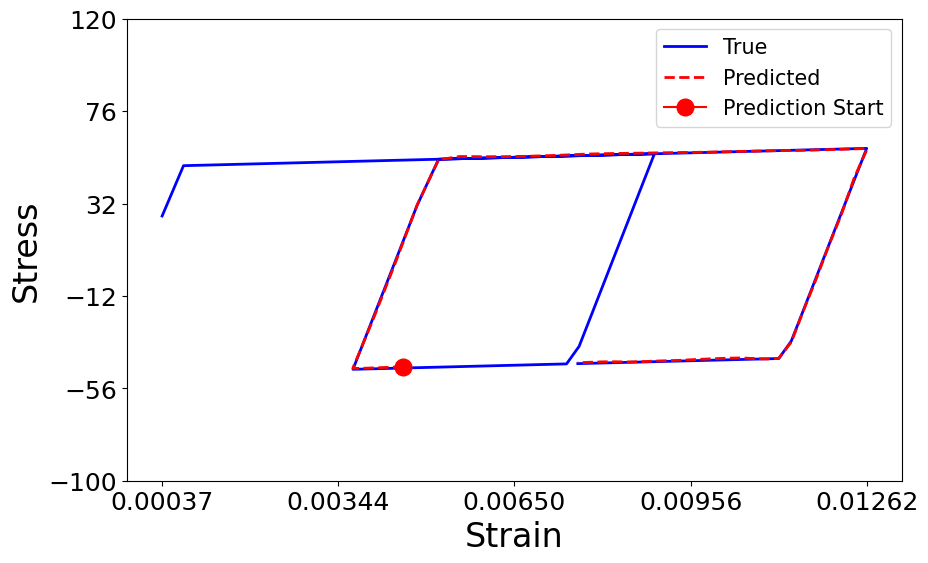}
        \caption{HANO prediction vs.~true response}
        \label{fig:HANO_incomplete}
    \end{subfigure}
    \hfill
    \begin{subfigure}[b]{0.32\textwidth}
        \centering
        \includegraphics[width=\textwidth]{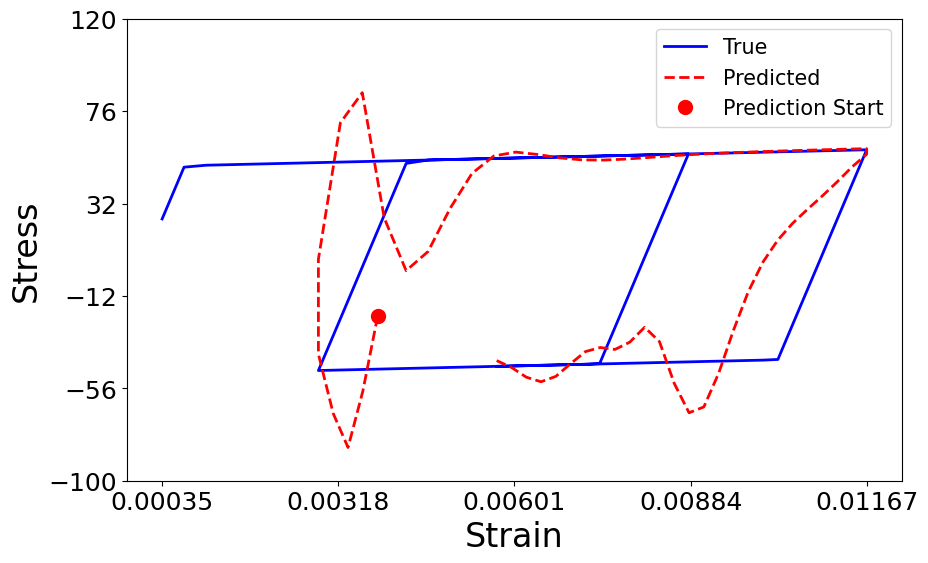}
        \caption{RNN$_1$ prediction vs.~true response}
        \label{fig:rnn1_incomplete}
    \end{subfigure}
    \hfill
    \begin{subfigure}[b]{0.32\textwidth}
        \centering
        \includegraphics[width=\textwidth]{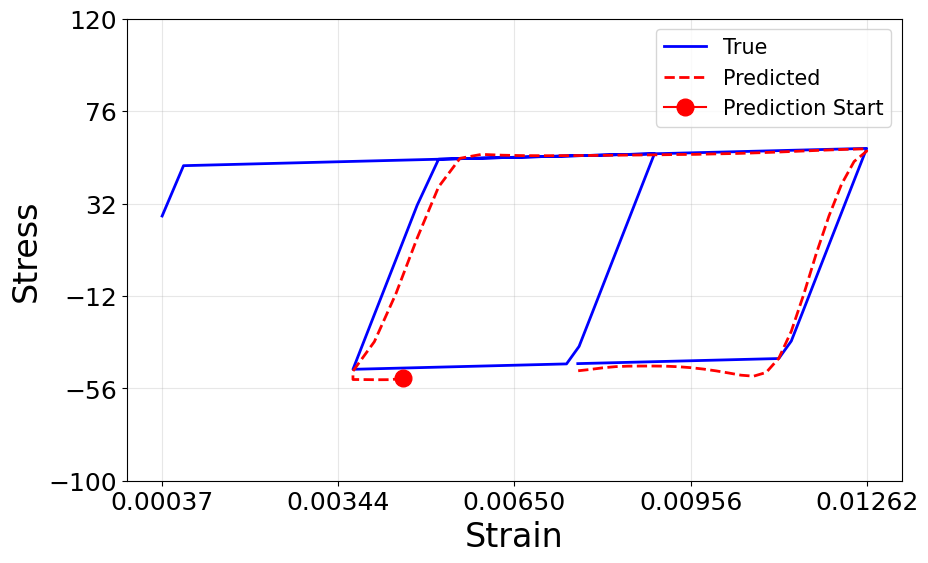}
        \caption{RNN$_2$ prediction vs.~true response}
        \label{fig:rnn2_incomplete}
    \end{subfigure}
    \caption{Comparison of model predictions on an exemplary case from Testset~II, where prediction starts at $45\,\%$ of the loading history. (a)~HANO; (b)~RNN$_1$; (c)~RNN$_2$. The red dot marks the deformed state at which prediction begins.}
    \label{fig:incomplete_history_prediction}
\end{figure}

Figure~\ref{fig:incomplete_history_prediction} compares  
HANO and RNN\(_1\) for a case selected from Testset II,
where approximately 45\% of the initial loading history was removed. 
HANO accurately predicts the subsequent material responses, closely matching the ground-truth.
The stress prediction by RNN\(_1\) exhibits significant  deviation 
from the reference immediately after forecasting begins, due to a mismatch in the  initialized hidden states.
This highlights RNN’s strong reliance on complete loading paths that closely resemble those seen during training.
Introducing stress history in RNN$_2$ helps alleviate the initial bias, yet the recurrent formulation still accumulates appreciable error during subsequent stress predictions. 
These observations reaffirm that the hidden state dynamics intrinsic to standard RNNs, rather than the input representation alone, limit their reliability for path‐dependent plasticity problems.

By incorporating both strain and stress information and employing the designed neural operator framework, the proposed HANO approach effectively overcomes the zero-stress initialization constraints that limit conventional RNN-based models.
This suggests its superior capacity for learning the hidden dynamics that govern path-dependent, irreversible behavior, enabling accurate stress predictions from any intermediate point along the loading history, even when initial data segments are missing. \textcolor{blue}{While this study primarily focuses on demonstrating HANO's ability to address methodological limitations of existing approaches , we also conduct parameter-matched comparisons with RNN-based models to ensure fair evaluation, as detailed in \ref{appx:parameter_matched_rnn}. Even under matched parameter counts, HANO demonstrates superior performance on both test scenarios, validating the effectiveness of the proposed architectural design.}

\subsection{Verification test: Discretization sensitivity}
\label{dis_invariance}
Another well-known issue  with RNN-based models is their lack of self-consistency (Limitation 1 in Section \ref{sec:pitfalls}), 
manifesting as sensitivity to the discretization of the loading path~\cite{bonatti2022importance}. 
This dependence on the resolution of loading steps violates the fundamental principle in mechanics that material constitutive (e.g., quasi-static) responses should depend solely on the loading path itself, not on how that path is sampled. In this test, we demonstrate the ability of the HANO approach to produce consistent predictions across varying discretization resolutions.

\subsubsection{Numerical test setup}
We employ the training dataset introduced in Section~\ref{sec:data_elastoplast}, in which each loading cycle is uniformly sampled with 100 increments.  
To investigate the discretization sensitivity, we further generate a family of test sets (Testset I)
that retain the same material parameters and strain–amplitude distributions 
but vary \emph{only}  in temporal/sampling resolution, spanning 60–150 increments per cycle.  

To ensure a fair comparison, we use RNN\(_2\) as a benchmark, whose input is the same concatenated strain–stress history window and forthcoming strain increment as in HANO.  
Both models are trained under the identical optimization protocols, using progressive scheduled sampling and epoch‑dependent Gaussian noise injection (see Section \ref{sec:training_schedule}).

\subsubsection{Effect of discretization resolution in testset}
\label{sec:resolution_testset}

\begin{figure}[htbp]
    \centering
    \includegraphics[width=0.85\textwidth]{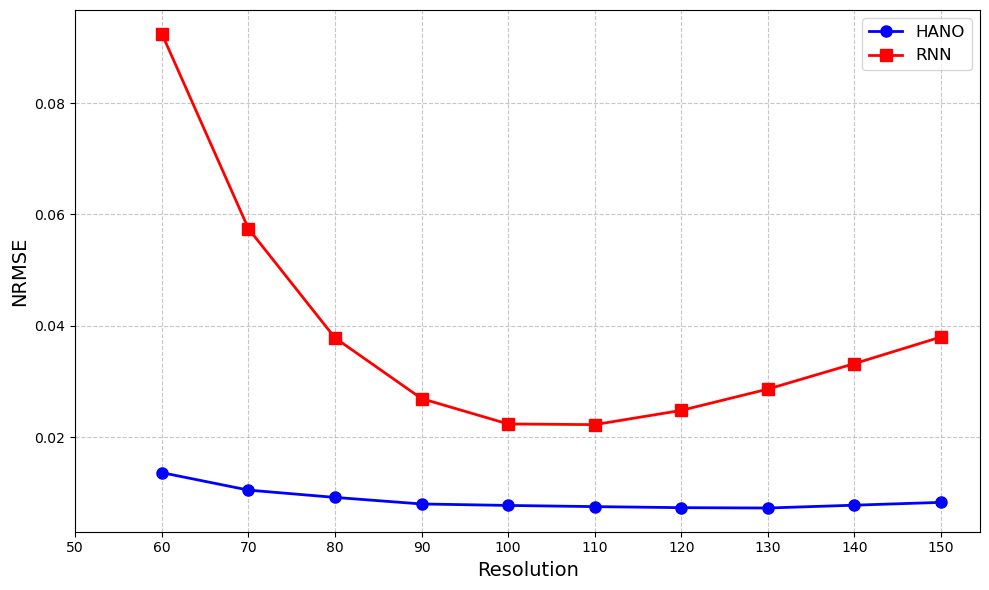}
    \caption{Comparison of prediction errors (NRMSE) for RNN\(_2\) and HANO approaches  with  varying discretization resolutions \textcolor{blue}{in test dataset}. Both models were trained on data with 100 time steps per loading cycle. The RNN model exhibits strong resolution dependence, with prediction error increasing as resolution deviates from the training resolution. In contrast,  HANO maintains consistently low errors across all  tested resolutions.}
    \label{fig:discretization_comparison}
\end{figure}

Figure~\ref{fig:discretization_comparison} presents the comparison of the HANO and RNN\(_2\) predictions on test datasets with varying discretization resolutions.  
In addition to HANO consistently reducing errors by nearly an order of magnitude (NRMSE \(0.6\%\) at \(100\) steps versus \(2\%\) for RNN\(_2\)), the results reveal a clear distinction in behavior between these two models.  
The RNN model exhibits strong resolution dependence, with prediction error increasing rapidly as the testing resolution deviates from the training setup of 100 time steps; specifically, its NRMSE rises to \(9\%\) at \(60\) steps and \(4\%\) at \(150\) steps.  
As expected, it achieves its highest accuracy when the testing resolution matches the training resolution.  
We also note that coarser resolutions (\(60\) steps) lead to larger errors than finer ones (\(150\) steps) relative to the training baseline.  

In contrast, the proposed HANO model maintains consistently low error across all resolution levels—\(1.1\%\) at \(50\) steps, \(1.2\%\) at \(60\) steps, \(0.6\%\) at \(100\) steps, and \(0.7\%\) at \(150\) steps—with only minimal degradation at the lowest resolutions.  
This highlights the resolution-invariant property of the HANO framework.  
Even at the lowest tested resolution (\(60\) time steps), HANO achieves better performance than the RNN model at its optimal resolution (\(100\) steps), with errors of \(1.1\%\) versus \(2\%\), demonstrating robustness to discretization variation that derives from its neural-operator backbone.  
The slight increase in HANO error at very coarse resolutions is attributed to path discontinuities that arise when the loading history is too sparsely sampled, introducing higher abrupt transitions that can challenge even resolution-invariant surrogates.

\begin{figure}[htbp]
    \centering
    \begin{subfigure}[b]{\textwidth}
        \centering
        \includegraphics[width=0.7\textwidth]{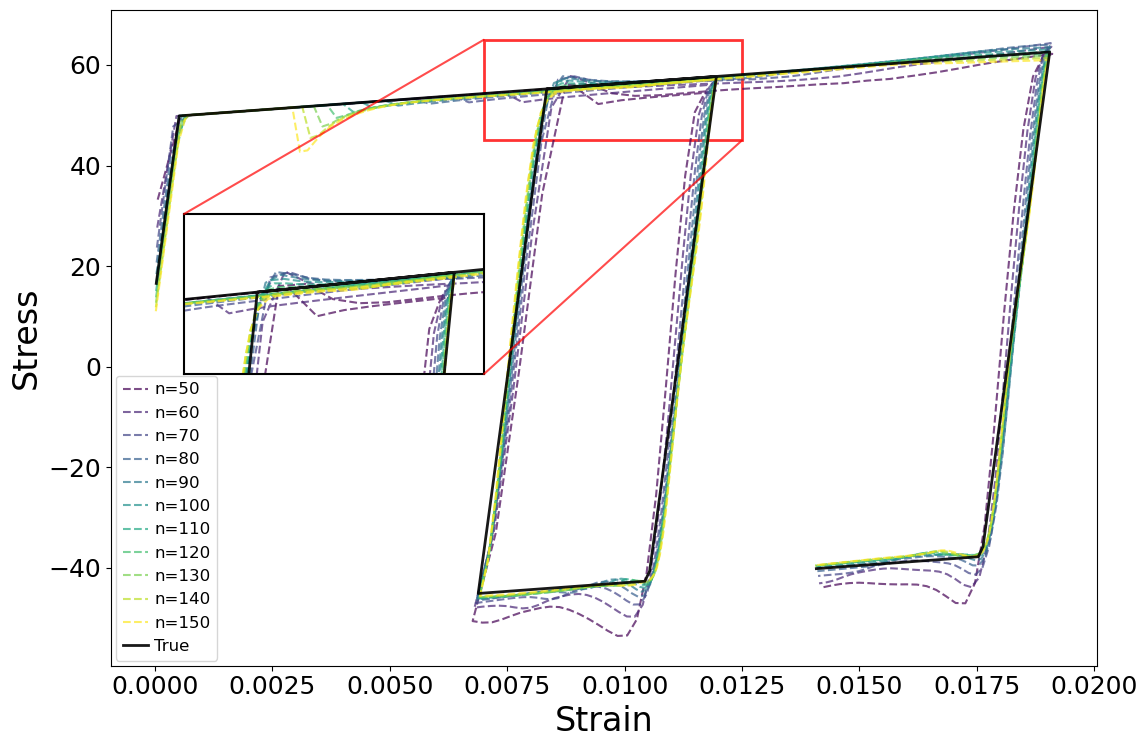}
        \caption{RNN\(_2\) predictions}
        \label{fig:rnn_resolution}
    \end{subfigure}
    \vspace{0.5cm}
    \begin{subfigure}[b]{\textwidth}
        \centering
        \includegraphics[width=0.7\textwidth]{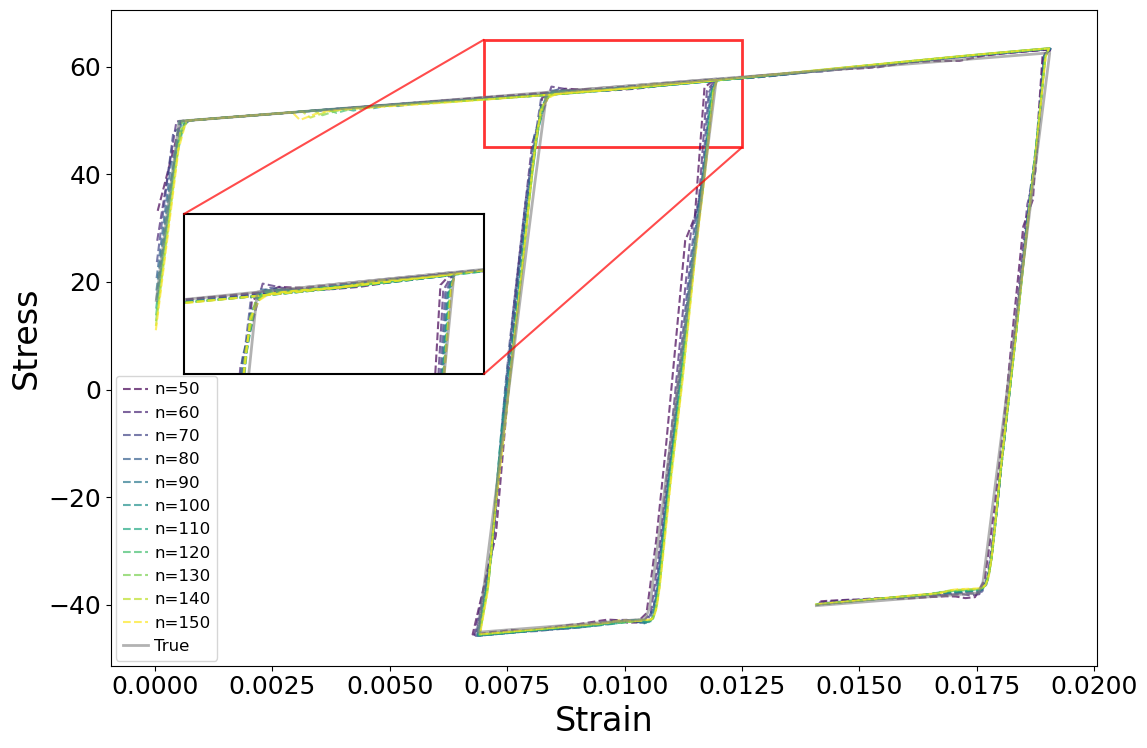}
        \caption{HANO predictions}
        \label{fig:HANO_resolution}
    \end{subfigure}
    \caption{Comparison of model predictions on  a representative test loading path across different temporal resolutions, where $n$ denotes the number of loading steps. (a)  RNN model shows progressively degraded accuracy as the resolution deviates from the training resolution. (b)  HANO model maintains consistent predictive accuracy across all resolutions.}
    \label{fig:resolution_predictions}
\end{figure}

To highlight their prediction performance at the loading-unloading transition areas, Figure~\ref{fig:resolution_predictions} shows predictions from both models on the same loading path sampled at different resolutions. 
In particular, RNN exhibits relatively large errors at sharp changes in the loading curves, especially at transitions between elastic and plastic behavior. 
On the other hand, HANO consistently captures these sharp transitions.
This  reveals that HANO can learn the underlying continuous dynamics mapping for inelastic material prediction, rather than merely fitting to a specific discretization scheme.

\subsubsection{Effect of discretization resolution in training data}
\label{sec:resolution_training}
To further validate our approach's discretization invariance, we conducted an additional experiment by training HANO on a dataset with varying resolutions. The training dataset contained 1000 loading cases with resolutions randomly distributed between 50 and 150 time steps per cycle, while maintaining the same material parameters and strain amplitude ranges as in previous experiments.

\begin{figure}[htbp]
    \centering
    \includegraphics[width=0.7\textwidth]{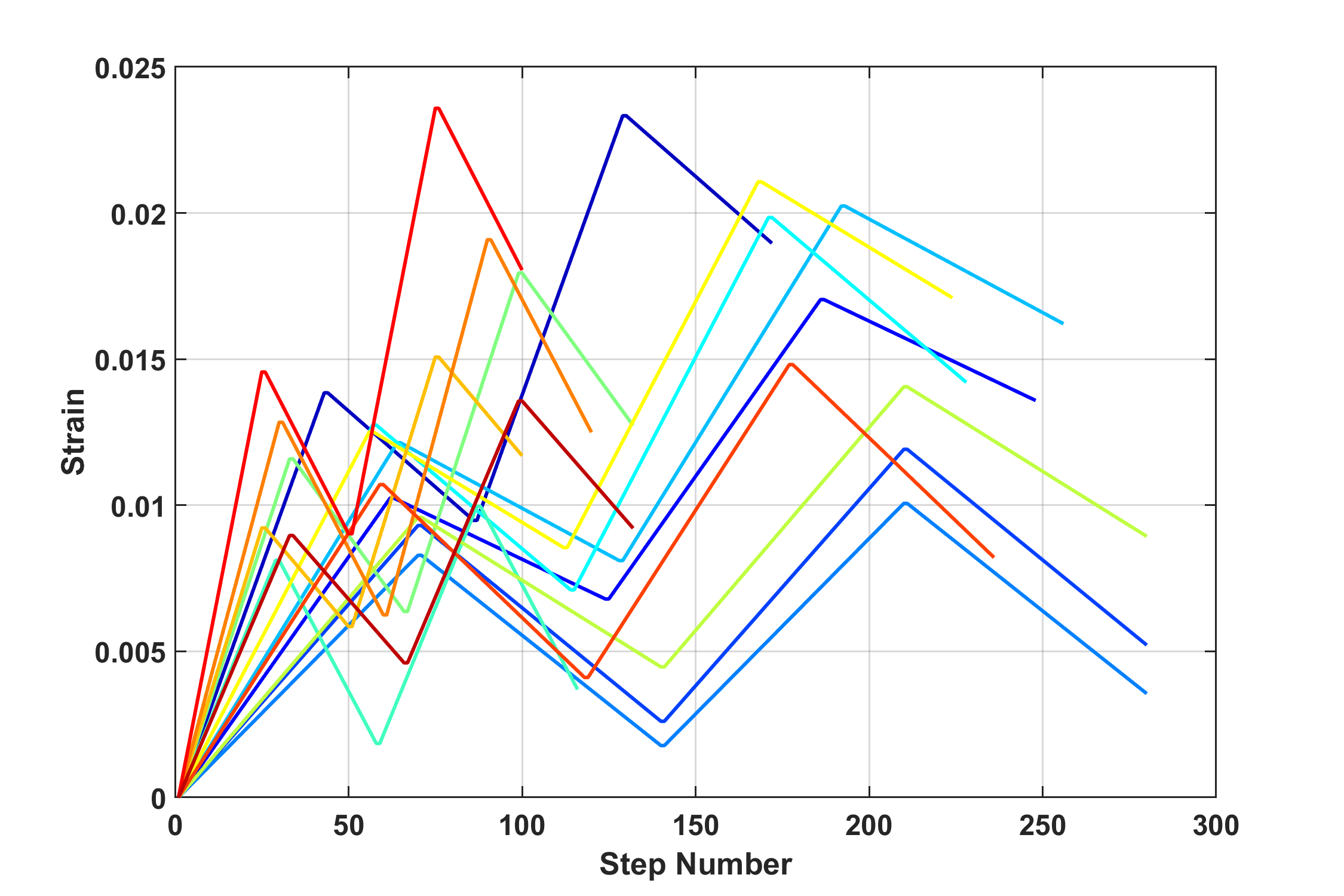}
    \caption{Representative strain loading paths from the variable-resolution training dataset, illustrating diversity in both sampling resolution and peak strain magnitude.}
    \label{fig:variable_resolution_samples}
\end{figure}

Figure~\ref{fig:variable_resolution_samples} presents representative strain loading paths from the training dataset,  each sampled at a different temporal resolution.
While all curves share a triangular loading–unloading structure, they vary significantly in both the number of steps (ranging from approximately 50 to 300) and peak strain amplitudes (from roughly 0.015 to 0.025). 
This variability exposes the model to a wide range of discretization granularities during training, equipping it to generalize across both coarse and fine temporal grids.

\begin{figure}[htbp]
    \centering
    \begin{subfigure}[b]{0.48\textwidth}
        \centering
        \includegraphics[width=\textwidth]{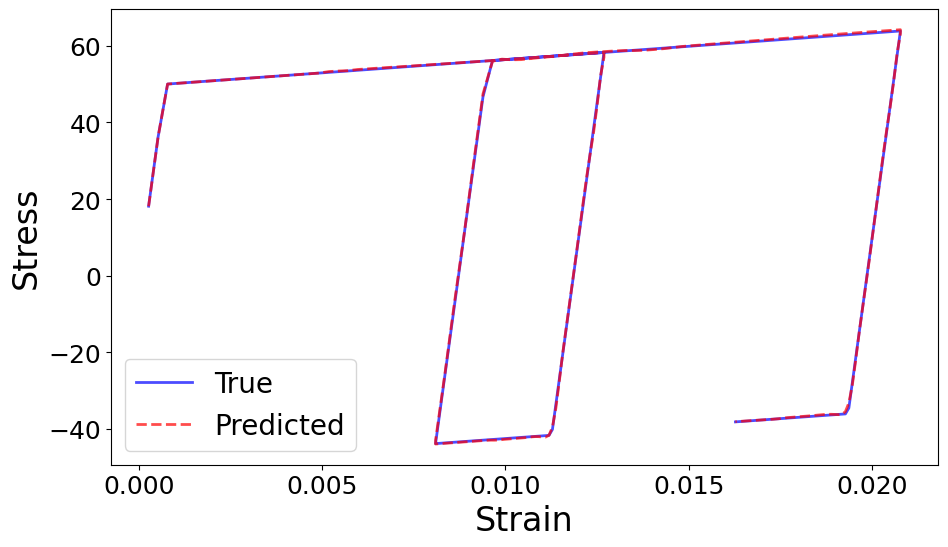}
        \caption{Resolution: 57 steps}
        \label{fig:test_case_1}
    \end{subfigure}
    \hfill
    \begin{subfigure}[b]{0.48\textwidth}
        \centering
        \includegraphics[width=\textwidth]{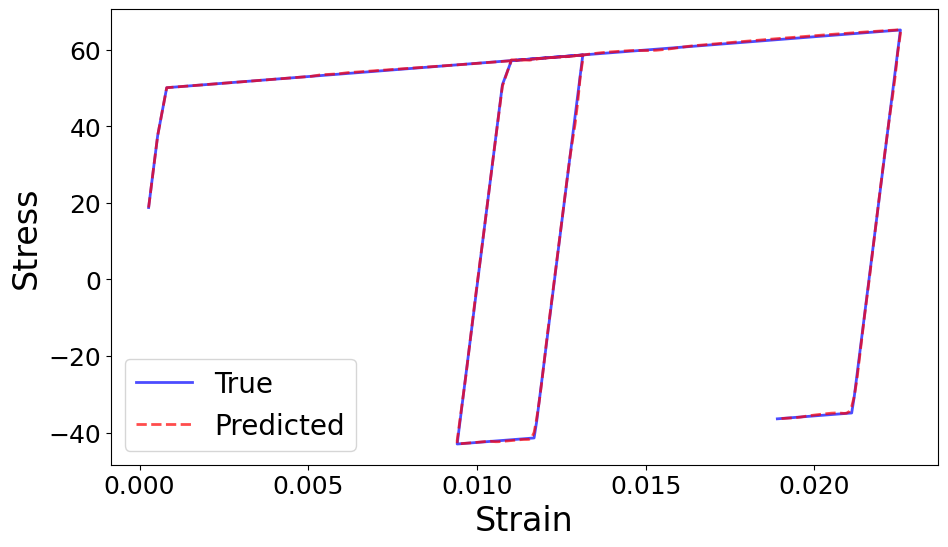}
        \caption{Resolution: 83 steps}
        \label{fig:test_case_2}
    \end{subfigure}
    
    \vspace{0.4cm}
    
    \begin{subfigure}[b]{0.48\textwidth}
        \centering
        \includegraphics[width=\textwidth]{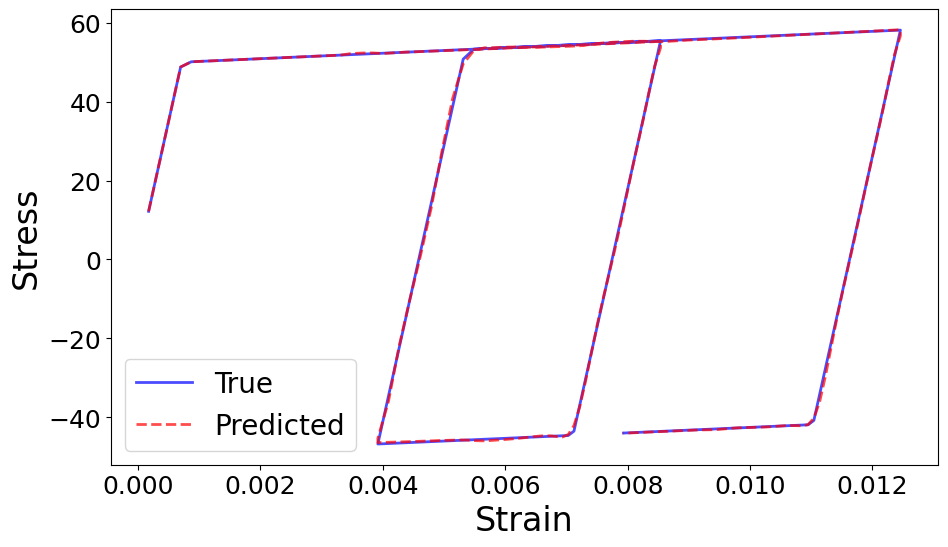}
        \caption{Resolution: 112 steps}
        \label{fig:test_case_3}
    \end{subfigure}
    \hfill
    \begin{subfigure}[b]{0.48\textwidth}
        \centering
        \includegraphics[width=\textwidth]{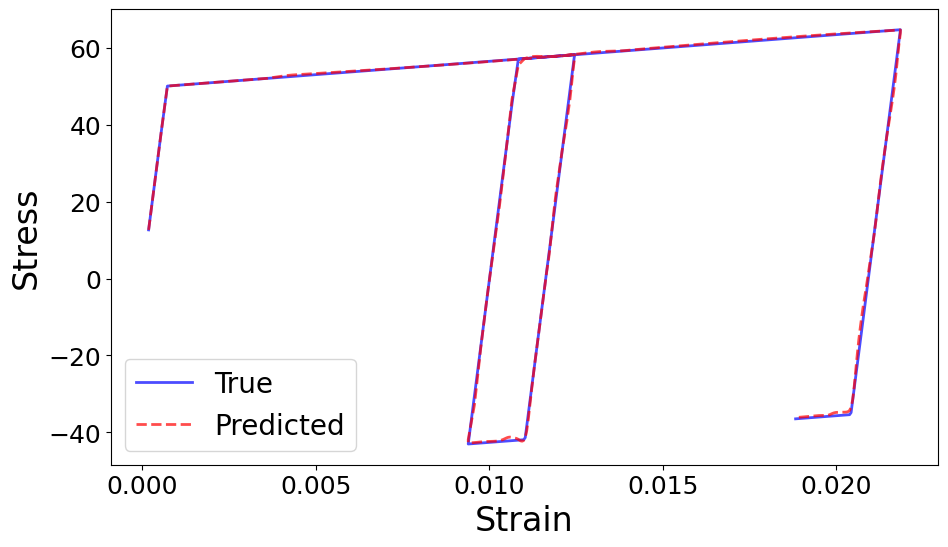}
        \caption{Resolution: 135 steps}
        \label{fig:test_case_4}
    \end{subfigure}
    
    \caption{Four representative test cases showing HANO predictions  compared to ground truth  across resolutions ranging from 57 to 135 time steps per loading cycle.}
    \label{fig:variable_resolution_examples}
\end{figure}

We then evaluated this model on an unseen test set with similar properties—random loading paths with randomly assigned resolutions. The retrained model attains an NRMSE of 0.6\%—essentially the same as the best single-resolution model in Section \ref{sec:resolution_testset}—demonstrating its ability to generalize across arbitrary discretization schemes without compromising accuracy. Figure \ref{fig:variable_resolution_examples} plots four representative cases covering 57, 83, 112, and 135 increments per cycle. In every panel, the purple HANO trace overlays the black reference curve so closely that the two are visually indistinguishable; even at the coarsest 57-step discretisation, yield turn-points, unloading slopes, and reloading plateaux align without noticeable 
deviations. 
This consistent performance across varying resolutions further validates that the proposed neural operator accurately captures  the underlying continuous mapping between strain and stress functions rather than becoming dependent on a specific discretization scheme.

In this section, the results clearly demonstrate that  HANO effectively resolves the self-consistency issue by learning mappings over continuous function spaces rather than discrete sequences. This capability makes HANO inherently more robust for practical applications where data may be sampled at irregular or varying rates. Given its consistent superior performance, we focus exclusively on HANO in the following discussions.

\textcolor{blue}{
To assess the training and convergence behaviors of the proposed HANO framework, we provide the corresponding training and validation loss curves in \ref{appx: appendix_loss_curve}.}
\subsection{Generalization under multiple loading cycles}
\label{sec:multi_cycle_extrapolation}
\textcolor{blue}{In this section, we assess the HANO model's ability to extrapolate beyond its training range. As shown in Table \ref{tab:dataset_1d}, the baseline HANO model was trained on the original fixed-resolution dataset from Section 4.1 (100 time steps per cycle, 2 loading cycles). To evaluate its generalization capability, we tested additional scenarios involving three, four, and five cycles.}

\begin{table}[htb]
    \centering
    \begin{tabular}{cc}
    \toprule
    \textbf{Number of Cycles (Test)} & \textbf{NRMSE} \\
    \midrule
    3 cycles & 0.006 \\
    4 cycles & 0.008 \\
    5 cycles & 0.008 \\
    \bottomrule
    \end{tabular}
    \caption{NRMSE results of the HANO model under different numbers of loading--unloading cycles}
    \label{tab:multi_cycle_nrmse}
\end{table}

\begin{figure}[htbp]
    \centering
    \includegraphics[width=0.7\textwidth]{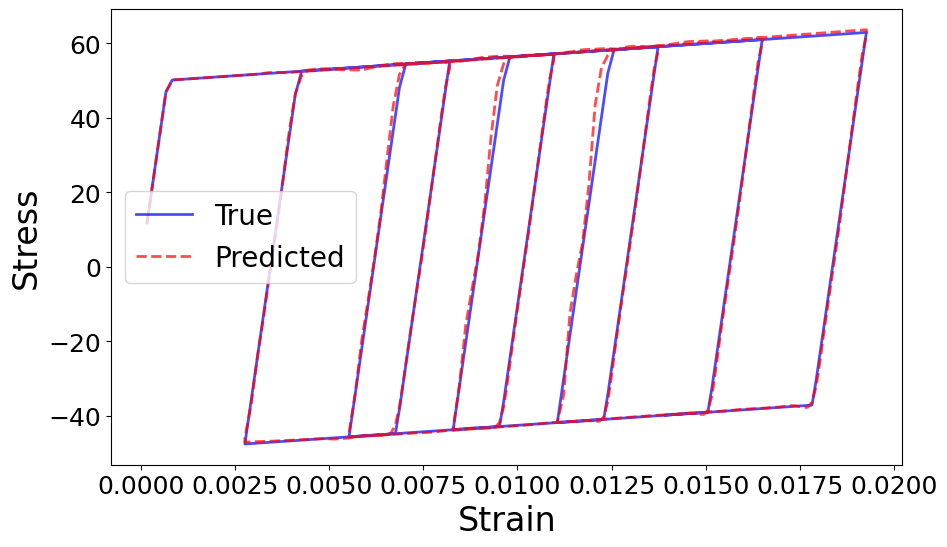}
    \caption{Comparison of predicted and true stress--strain responses over five complete loading--unloading cycles. }
    \label{fig:five_cycle_example}
\end{figure}

Table~\ref{tab:multi_cycle_nrmse} shows that the loss remains small for every
test case.  Even at five cycles the NRMSE rises from the two-cycle reference in
Section~\ref{sec:ref_1d} by only \(0.2\%\) (from \(0.006\) to \(0.008\)),
demonstrating that the operator keeps nearly the same accuracy as the cycle
count increases.
Figure~\ref{fig:five_cycle_example} reveals that HANO tracks the reference response with only a very small deviation at every yield point, reverse-yield point, and hysteresis loop area, and no cumulative drift is observed.
The hardening–unloading–rehardening slope remains unchanged from one cycle to the next, demonstrating that the model faithfully reproduces the full evolution of the material’s hardening behaviour throughout repeated loading.
Taken together, these findings confirm that HANO stays stable over long loading histories: after each reversal the operator window promptly realigns with the current hardening behaviour instead of merely replaying the patterns present in the training data.

\subsection{\textcolor{blue}{Robustness to noisy initial condition}}
\label{sec:noise_robustness}
Real experimental measurements of strain and stress are inevitably affected by sensor noise. To verify whether the proposed HANO method can effectively filter out such noise and maintain accurate plasticity predictions, we conducted an experiment by superimposing random noise onto the \textcolor{blue}{initial stress signals} and comparing the resulting predictions against the noise-free ground truth. 

Following the elasto-plastic setup in Section~\ref{sec:data_elastoplast}, we added Gaussian noise to the $k$ initial stress inputs:
\[
    \sigma^\mathrm{(noisy)}_t
    \;=\; 
    \sigma^\mathrm{(true)}_t
    \;+\;
    \mathcal{N}\bigl(0,\;(\alpha \,\sigma_\mathrm{std})^2\bigr),
\]
where $\alpha = 10\%$ is the noise ratio, and $\sigma_\mathrm{std}$ denotes the standard deviation of the underlying noise-free stress signals. After injecting this noise, we used the resulting \emph{noisy} stress as part of the input to HANO, and quantified the prediction errors with respect to the noise-free reference.
\textcolor{blue}{We note that due to HANO's autoregressive prediction mechanism, noise is only added to the stress components within the initial history window. Once HANO begins making predictions, subsequent stresses are generated by the model itself rather than being provided as external data.}

\begin{figure}[htbp]
    \centering
    \includegraphics[width=0.75\textwidth]{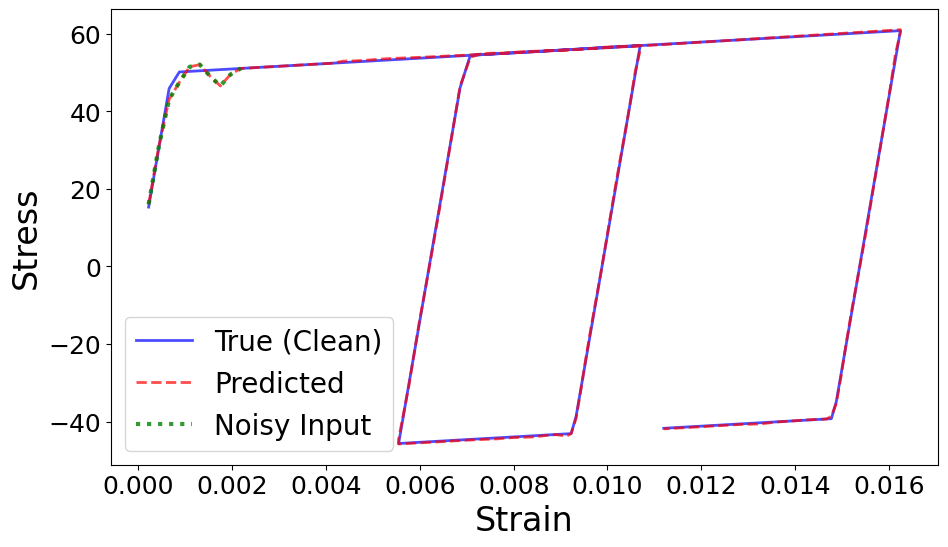}
    \caption{Example of the noisy stress input ($\alpha=10\%$), HANO prediction, and the true (noise-free) stress. The model successfully filters out random fluctuations and maintains close agreement with the underlying material response.}
    \label{fig:HANO_noise_comparison}
\end{figure}

Figure~\ref{fig:HANO_noise_comparison} shows a representative stress-strain path with noisy input (using $\alpha = 10\%$), the HANO prediction, and the corresponding noise-free
reference curve.  As seen in the figure, the high-frequency components of the
noise are strongly suppressed, and the predicted curve runs smoothly through
the centre of the noisy envelope.  
This behaviour reflects the operator’s
projection effect: the network maps the noisy input back onto the learned
physical response manifold, acting as an implicit filter.  Quantitatively, the
NRMSE is around 0.015, demonstrating only a mild reduction in accuracy
compared to the noiseless case.

Overall, these findings suggest that HANO is inherently robust to initial sensor noise: it learns an operator mapping that respects the governing material laws, rather than overfitting to noisy observations. In practical applications where imperfect sensor data are unavoidable, this robustness ensures reliable stress predictions from potentially fluctuating signals.

%% file: Numerical_damage.tex
\section{Anisotropic Damage in Brittle Solids}\label{sec:damage}
In this section, we examine the predictive capacity of the proposed data-driven constitutive model, HANO, for brittle solids undergoing progressive anisotropic damage. 
Specifically, we conduct comparative experiments to assess how different configurations of the HANO model affect the accuracy of stress predictions under various path-dependent loading conditions,
considering factors 
such as the integration of U-Net architectures, the placement of attention component, and the number of history input steps.

\subsection{Benchmark dataset for anisotropic damage}
\label{sec:data:anisodamage}
In this study, we use the open-access dataset released by~\citet{ge2024data}. The data are generated using \textsc{Abaqus/Explicit} simulations of a single eight-node brick element (type \texttt{C3D8R}) implemented with the modified Hashin damage model \cite{hashin1980failure}. 
This model effectively captures key failure mechanisms in unidirectional fiber-reinforced polymers (FRPs), including fiber tearing, matrix cracking, and fiber–matrix interface debonding.

The generation of loading–unloading trajectories is briefly described as follows. Each simulation prescribes a pseudo-random six-dimensional strain path,
\[
  \boldsymbol{\varepsilon}
  = \bigl[ \varepsilon_{xx}, \varepsilon_{yy}, \varepsilon_{zz},
            \gamma_{xy}, \gamma_{xz}, \gamma_{yz} \bigr],
\]
which is  updated at every increment according to
\[
  \Delta\varepsilon_{i}
  =
  \operatorname{sgn}\bigl(r_{1}-\theta\bigr)
  \Bigl[
    \Delta\varepsilon_{\min i}
    + r_{2}\bigl(\Delta\varepsilon_{\max i}-\Delta\varepsilon_{\min i}\bigr)
  \Bigr],
  \qquad
  r_{1},r_{2}\sim\mathcal{U}(0,1),
\]
where $r_{1}$ and $r_{2}$ are loading factors sampled randomly from the uniform distribution $\mathcal{U}(0,1)$, with parameters
\(\varepsilon_{\max}=0.05\),
\(\Delta\varepsilon_{\max}=0.01\),
\(\Delta\varepsilon_{\min}=0.005\),
and a symmetry parameter \(\theta=0.5\) that gives equal probability to
positive and negative increments.  
Loading terminates when
\(\|\boldsymbol{\varepsilon}\|_{2}\ge\varepsilon_{\max}\).
Since the constitutive law is rate-independent, the absolute time scale (and hence the nominal step size) has no influence on the stress–strain trajectories.

The published dataset by \citet{ge2024data} contains \(500\) independent trajectories,
each providing  the full six-component stress history corresponding to the imposed strain paths.
In the
present work, we use \(400\) for training and reserve \(100\) for testing.
Unless otherwise noted, all stress and strength values are reported in megapascal (MPa).  
For completeness, 
the governing equations describing the material model, along with 
the material parameters used in the original simulations are detailed in \ref{appx:hashin} and \ref{appx:params}, respectively.

Before evaluating model performance, we outline the proposed HANO architecture for clarity. 
The HANO model consists of six sequential operator layers,  comprising $L=3$ 
standard Fourier layers followed by $M=3$ advanced AEUF layers.
Each Fourier layer performs global spectral convolution with a spectral resolution of 10 Fourier modes (via FFT), and all layers utilize a channel width of $w=64$ for feature representations. 
For a fair comparison, all  HANO  variants (see Table \ref{tab:aufno_variants})  are implemented with the same spectral resolution (10 modes) and channel width (64), differing only in their layer structures. The input to HANO
consists of a window of $k=20$ previous strain–stress pairs concatenated with the next strain increment, and the network outputs the corresponding predicted stress for that increment.

\subsection{Study of attention placement in HANO architectures}
\label{sec:attention_placement}
\begin{figure}[!ht]

    \begin{subfigure}[b]{0.9\textwidth}
        \centering
        \includegraphics[width=\textwidth]{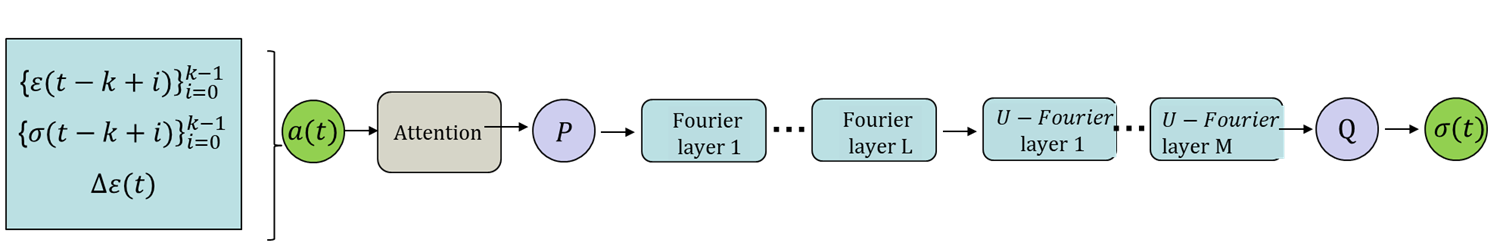}
        \caption{HANO$_1$: Attention mechanism applied to input}
    \end{subfigure}
  
    \begin{subfigure}[b]{0.9\textwidth}
        \centering
        \includegraphics[width=\textwidth]{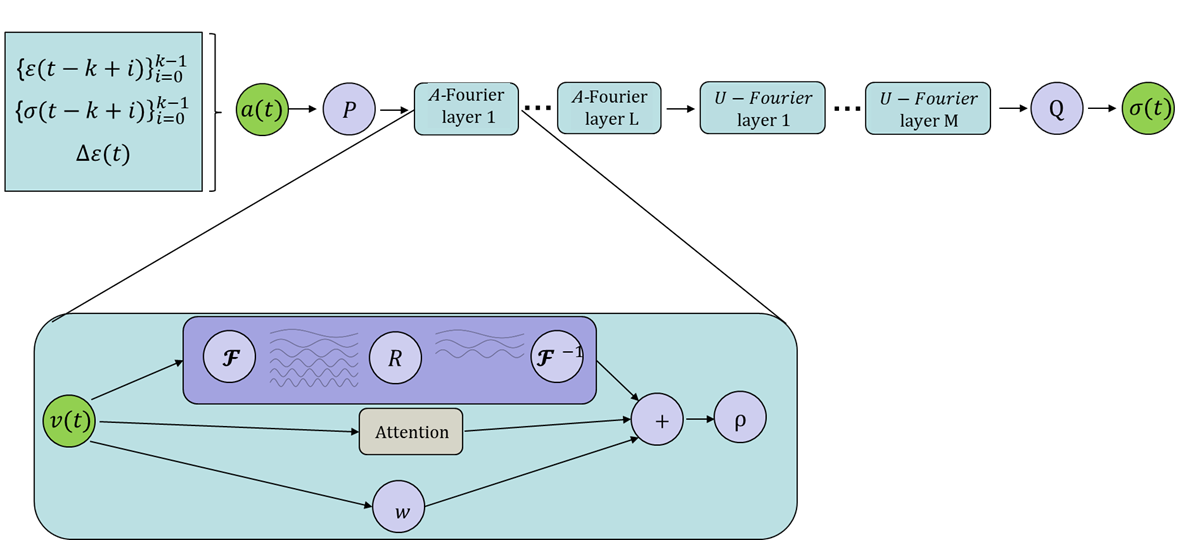}
        \caption{HANO$_2$: Attention mechanism parallel to Fourier layer}
    \end{subfigure}
    
    \vspace{1em}

    \vspace{1em}
    
    \begin{subfigure}[b]{0.9\textwidth}
        \centering
        \includegraphics[width=\textwidth]{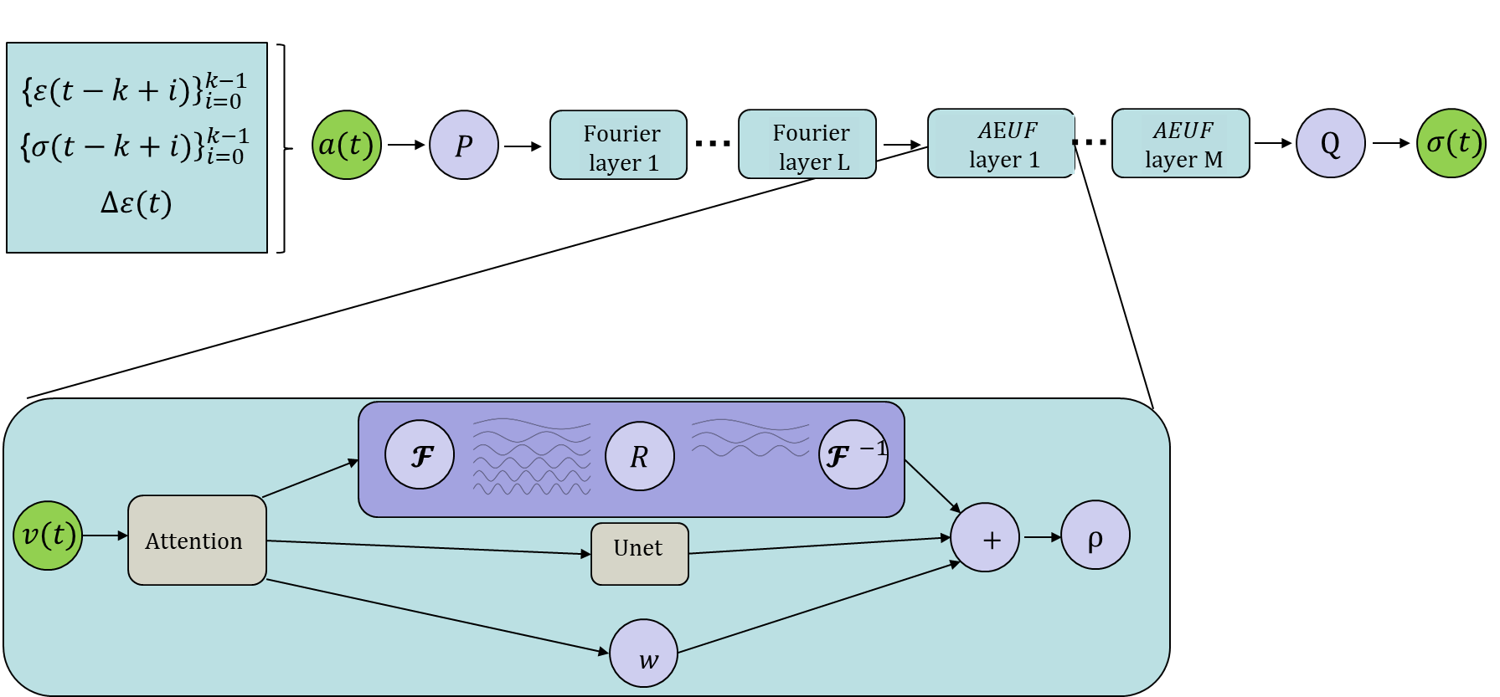}
        \caption{HANO$_3$: Attention mechanism applied before Fourier and U-Net layers}
        \label{fig:hano3}
    \end{subfigure}
    
    \caption{Architectures of three HANO variants with attention placed in different parts of the network.  
    (a) HANO$_1$ applies an attention mechanism to the input feature map before the Projection operator. 
    (b) HANO$_2$ integrates an attention module that works in parallel with the Fourier layer.   
    (c) HANO$_3$ incorporates attention mechanisms as pre-processing before both the Fourier and U-net layers.}
    \label{fig:model-comparison}
\end{figure}

To examine how the positioning of the self-attention module, as described in Section \ref{sec:attn}, influences predictive accuracy, we design three variants of HANO,
each distinguished by a different  strategy for integrating attention. 
Figure~\ref{fig:model-comparison} schematically illustrates these configurations.

The first variant, HANO$_1$, 
represents a naive introduction of self-attention, where the self-attention block is applied directly to the original input feature map. 
The intention here is to allow attention-driven refinements to guide the multi-scale feature extraction from the earliest stages of the network. 
The second variant, HANO$_2$, follows a structure similar to that proposed in~\citep{peng2022attention}, which shows improved prediction accuracy over the vanilla FNO in the context of spatial physics simulations.
In this setup, the self-attention block is applied to the feature map after the projection operator, and the resulting attention features are fused with outputs from both the Fourier layer and a linear transformation.
Lastly, HANO$_3$ introduces a self-attention module serially integrated with each U-Fourier layer, allowing both the downsampling and upsampling stages to benefit from attention-driven refinement of the feature maps. 
\textcolor{blue}{This third variant, HANO$_3$, serves as the default  architecture in this work, as illustrated in Figure \ref{fig:structure_model}.}

Table~\ref{tab:aufno_variants} compares the predictive performance of the three HANO variants against the baseline FNO in terms of NRMSE.  Among the models, HANO$_3$ achieves the highest accuracy, with an NRMSE of 21.09\%, representing a
40.1\%
reduction over FNO. 
Additionally, HANO$_2$, which incorporates multiple embedded attention modules,  yields roughly a 10\% error reduction relative to HANO$_1$.
These results suggest  that the repeated integration of attention mechanisms throughout the U-Net architecture   in HANO$_3$ 
enables more effective refinement of feature representations 
than a single attention block at the network input (HANO$_1$)
or 
parallel attention fusion 
with the Fourier layers before the U-Fourier layers (HANO$_2$). 

\begin{table}[!ht]
    \centering
    \begin{tabular}{|l|r|r|}
        \hline
        \textbf{Model} & \begin{tabular}[c]{@{}r@{}}\textbf{Performance}\\\textbf{(NRMSE)}\end{tabular} & \begin{tabular}[c]{@{}r@{}}\textbf{Improvement}\\\textbf{vs. FNO}\end{tabular} \\
        \hline
        HANO$_1$ & 26.53\% & 24.7\% \\
        \hline
        HANO$_2$ & 23.32\% & 33.8\% \\
        \hline
        HANO$_3$ & 21.09\% & 40.1\% \\
        \hline
    \end{tabular}
    \caption{Comparison of the three HANO variants in terms of predictive performance. }
    \label{tab:aufno_variants}
\end{table}

Overall, this analysis demonstrates that both the placement and integration strategy of attention mechanisms within the neural operator architecture significantly influence model performance.
Among the tested configurations, HANO$_3$ achieves more accurate stress predictions under complex loading histories.
While this study does not exhaustively explore all possible variants, the results underscore the promise of multi-stage self-attention for enhancing multiscale feature learning in path-dependent and anisotropic material systems.

\subsection{Comparison with Baseline Methods: FNO, UFNO, and HANO}
\label{sec:comparison_baseline}

Having established the importance of attention-placement strategies, we next compare the predictive performance of the proposed neural operator architecture against baseline models. Specifically, we contrast the standard Fourier Neural Operator (FNO), the U-Net-enhanced FNO (UFNO), and the top-performing  HANO variant, HANO$_3$ to illustrate how U-Net components and self-attention mechanisms improve stress predictions in history-dependent materials. For brevity,  we hereafter refer to  HANO$_3$ simply as HANO.
\textcolor{blue}{For comparison, the baseline models are configured as follows: FNO consists of 6 sequential Fourier layers with a channel width of $w = 64$ and spectral resolution of 10 Fourier modes. UFNO comprises 3 standard Fourier layers followed by 3 U-Fourier layers (each incorporating parallel U-Net pathways), also with $w = 64$ and 10 modes. Both baseline models use the same lifting and projection operators as HANO.}

Table~\ref{tab:model-comparison} presents the parameter counts, NRMSE values, and performance improvements of these three models. UFNO reduces the error of the baseline FNO by nearly 30\%, whereas HANO achieves an additional 15\% improvement over UFNO (and more than 40\% over FNO). These gains underscore the complementary benefits of U-Net–based multi-scale representations and self-attention–driven feature refinement. Notably, while HANO requires 22.6\% more parameters than FNO, the magnitude of the accuracy improvement indicates that the architectural enhancements—particularly the repeated application of attention within the U-Fourier layers—are the primary drivers of the performance gains. In other words, simply increasing the model size does not fully account for these improvements; rather, the carefully designed multi-scale and attention mechanisms play a decisive role in capturing history-dependent stress responses.

\begin{figure}[htbp]
\centering
\begin{subfigure}[b]{0.75\textwidth}
    \caption{}
    \centering
    \includegraphics[width=\textwidth]{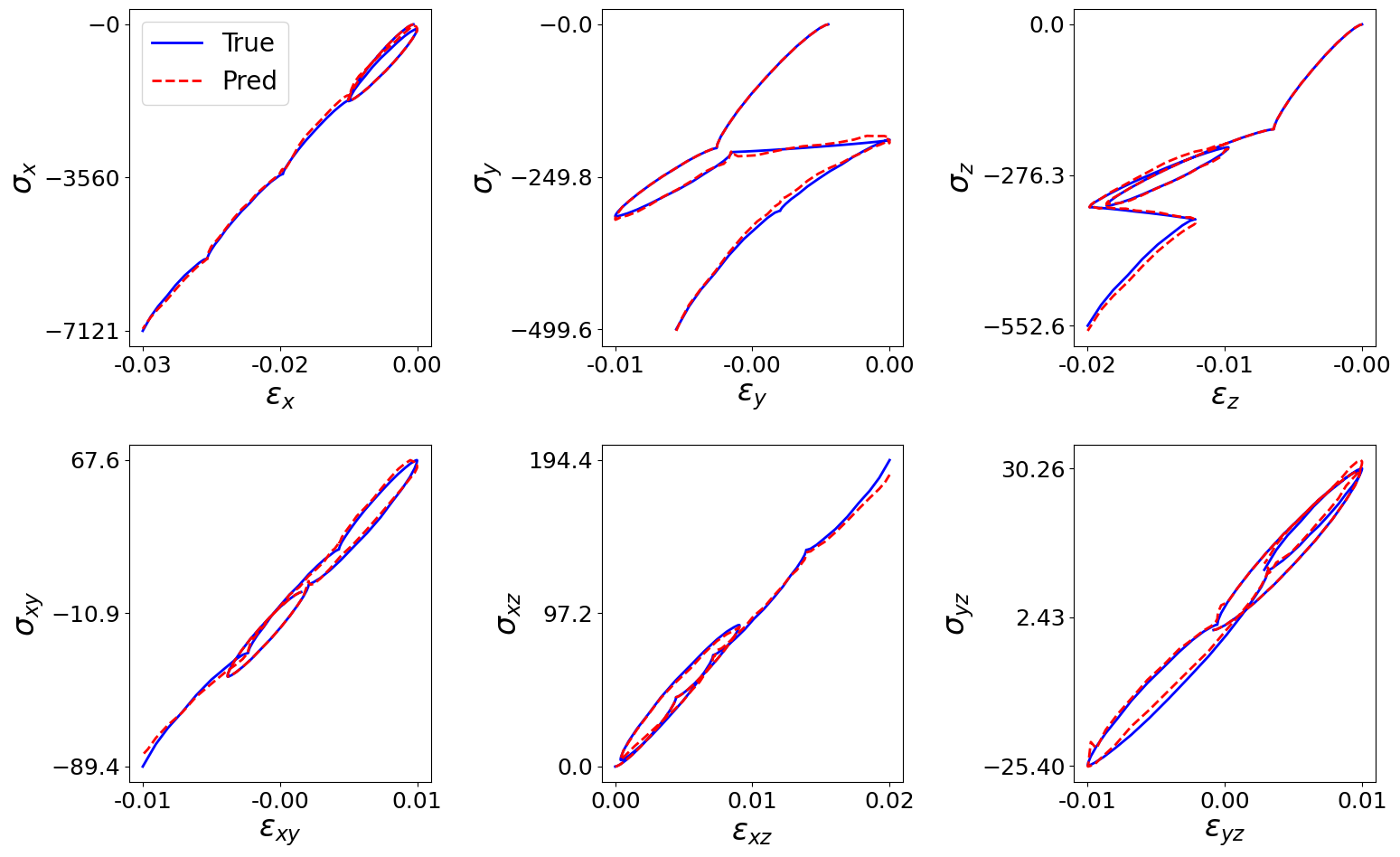}
    \label{fig:HANO}
\end{subfigure}

\begin{subfigure}[b]{0.75\textwidth}
    \caption{}
    \centering
    \includegraphics[width=\textwidth]{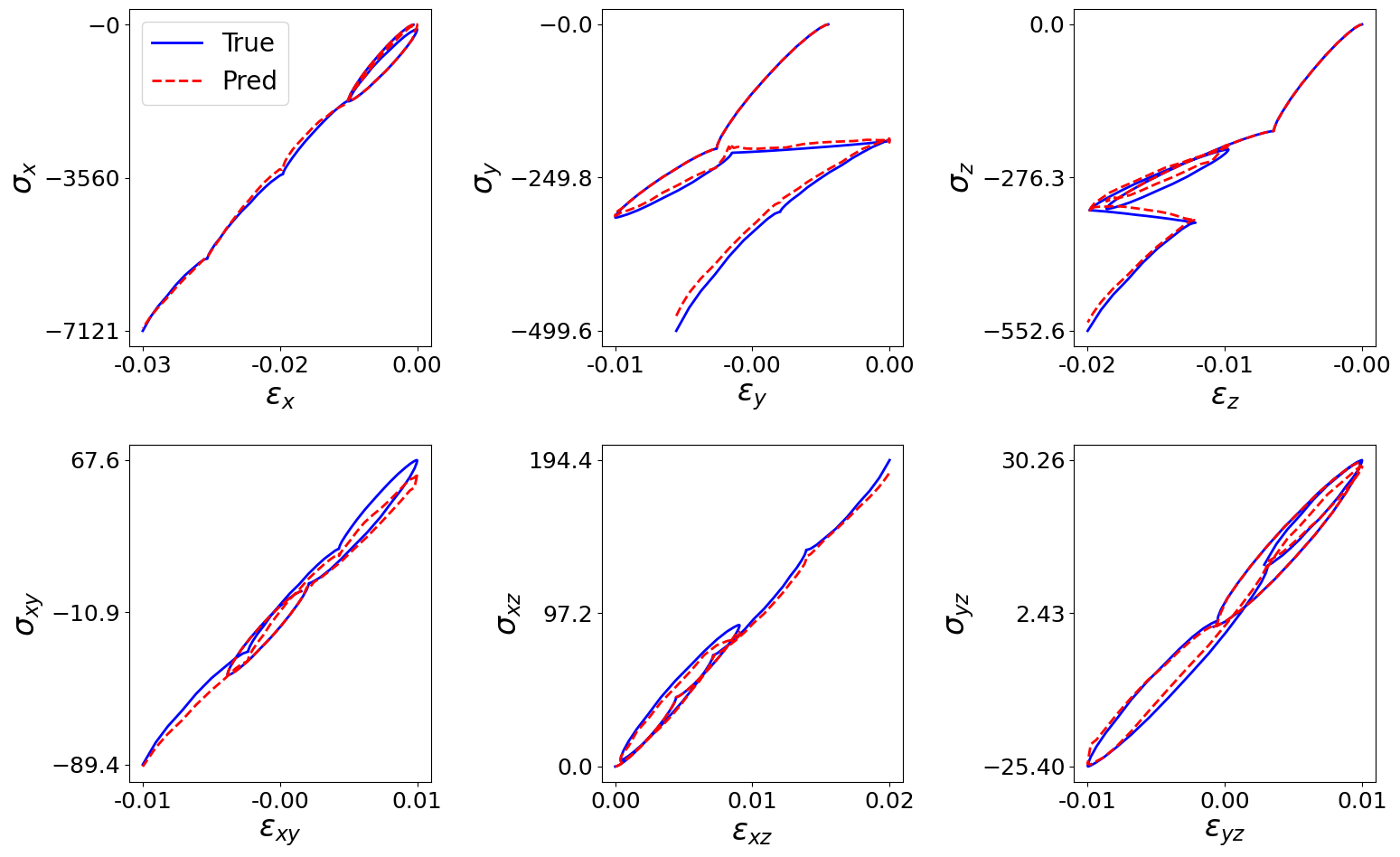}
    \label{fig:ufno}
\end{subfigure}

\begin{subfigure}[b]{0.75\textwidth}
    \caption{}
    \centering
    \includegraphics[width=\textwidth]{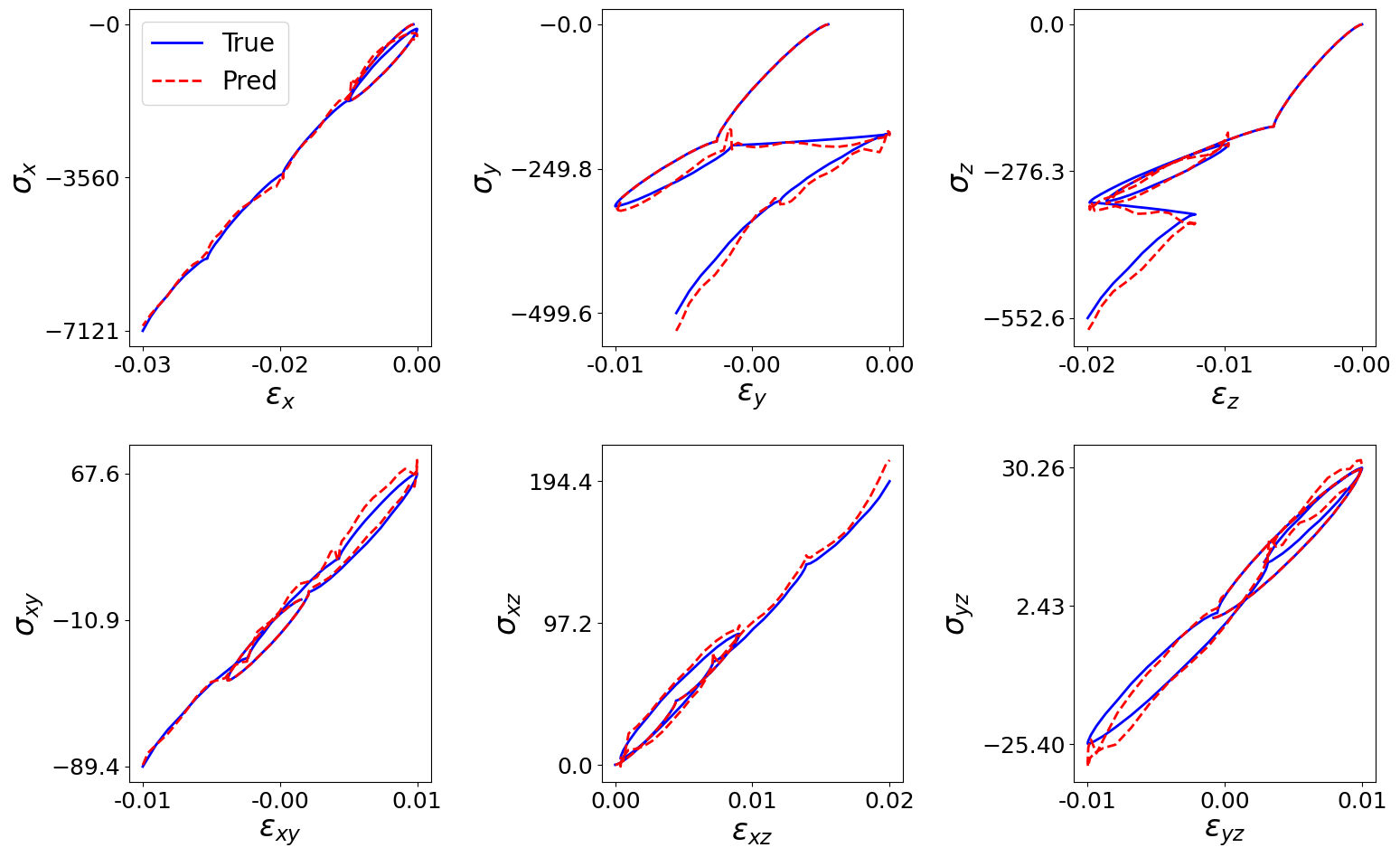}
    \label{fig:fno}
\end{subfigure}
\caption{Stress predictions using different neural operator architectures: (a) HANO, (b) UFNO, and (c) FNO.}
\label{fig:model_comparison}
\end{figure}

For a more intuitive illustration of the respective contributions of the U-Net and attention mechanisms, we further evaluate FNO, UFNO, and HANO on a single test case, as shown in Figure~\ref{fig:model_comparison}. The baseline FNO (Figure~\ref{fig:model_comparison}c) follows the stress curve well in smoothly varying regions, but it rounds off the yield and unloading corners and lags slightly at the peaks.
Introducing the U-Net component in UFNO markedly improves those sharp transitions: the multi-scale pathway preserves the corner angles, better captures the abrupt slope changes, and brings the peak stresses much closer to the reference.
The addition of  self-attention mechanism in HANO 
further improves the prediction by refining local stress features and mitigating long-term errors induced by complex strain–stress interactions. 
This improvement is particularly evident in the shear stress component  $\sigma_{xy}$, where both FNO and UFNO lose accuracy over several cycles, whereas HANO consistently maintains high fidelity.

\textcolor{blue}{To further evaluate the predictive performance of different neural operators with comparable model complexity, a parameter-matched comparison is conducted in Table~\ref{tab:parameter_matched} by enlarging the baseline models through increased channel widths: FNO-Large (width 72, $\sim$545k parameters) and UFNO-Large (width 68, $\sim$549k parameters). As shown in Table~\ref{tab:parameter_matched}, despite having similar parameter counts, HANO (21.09\% NRMSE) still outperforms both FNO-Large (33.76\%) and UFNO-Large (24.12\%). 
The enlarged FNO and UFNO models yield only marginal improvements, indicating that the performance gains of HANO primarily originate from the attention-enhanced U-Fourier architecture rather than from increased model size.
}

\begin{figure}[htbp]
    \centering
    \begin{subfigure}[t]{0.7\textwidth}
      \caption{Test case 1}
      \includegraphics[width=\textwidth]{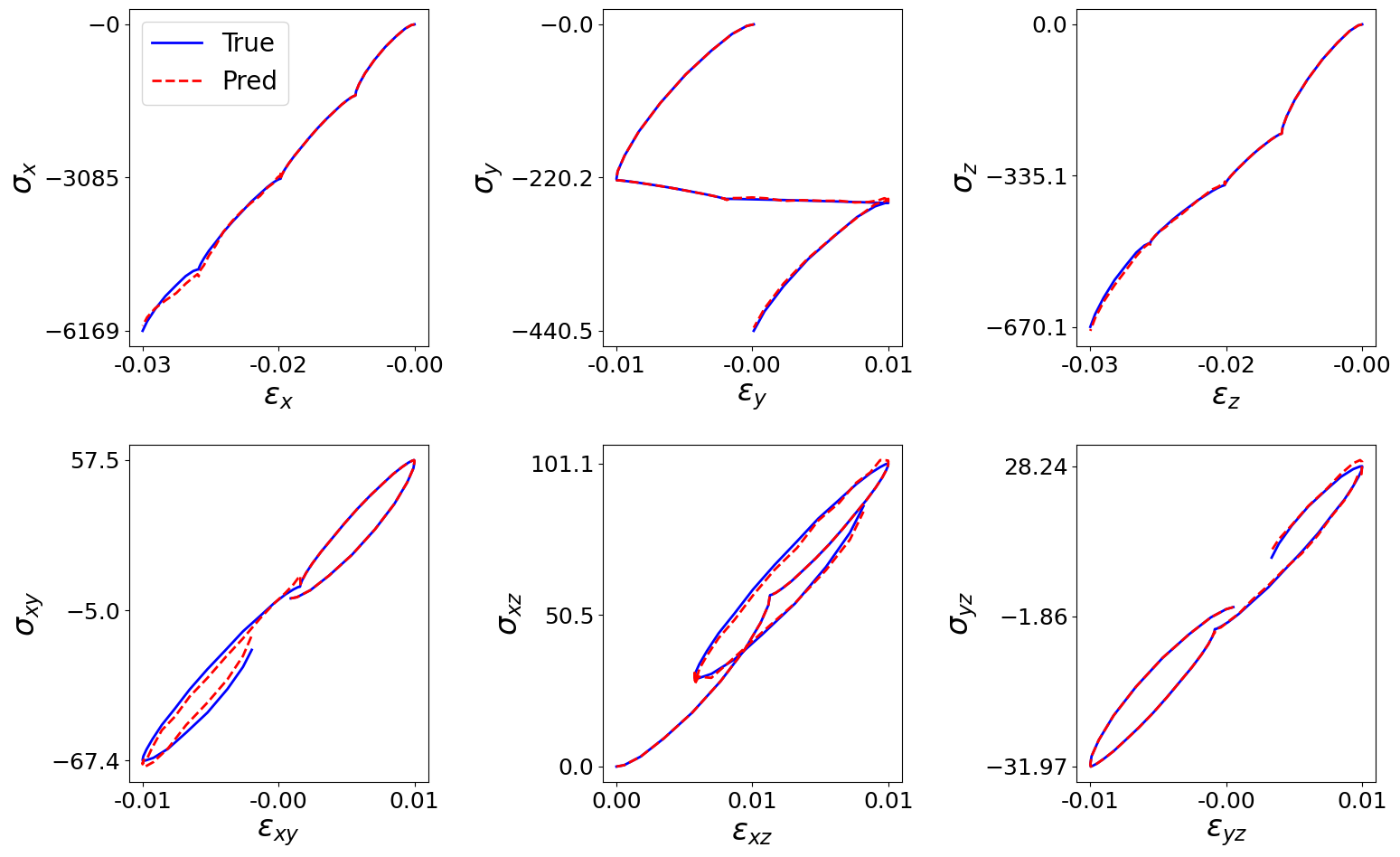}
      \label{fig:case1}
    \end{subfigure}

    \begin{subfigure}[t]{0.7\textwidth}
      \caption{Test case 2}
      \includegraphics[width=\textwidth]{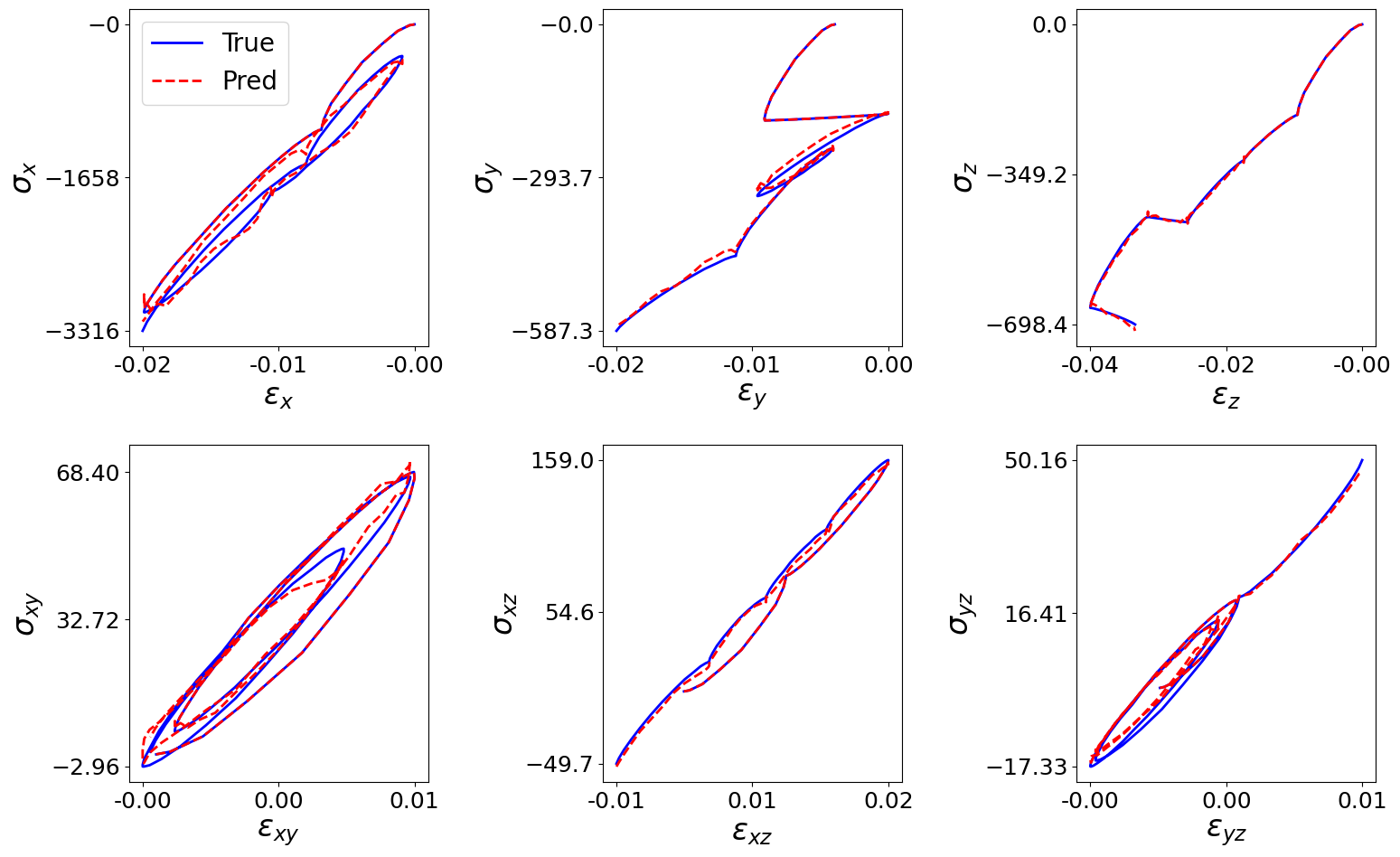}
      \label{fig:case2}
    \end{subfigure}

    \begin{subfigure}[t]{0.7\textwidth}
      \caption{Test case 3}
      \includegraphics[width=\textwidth]{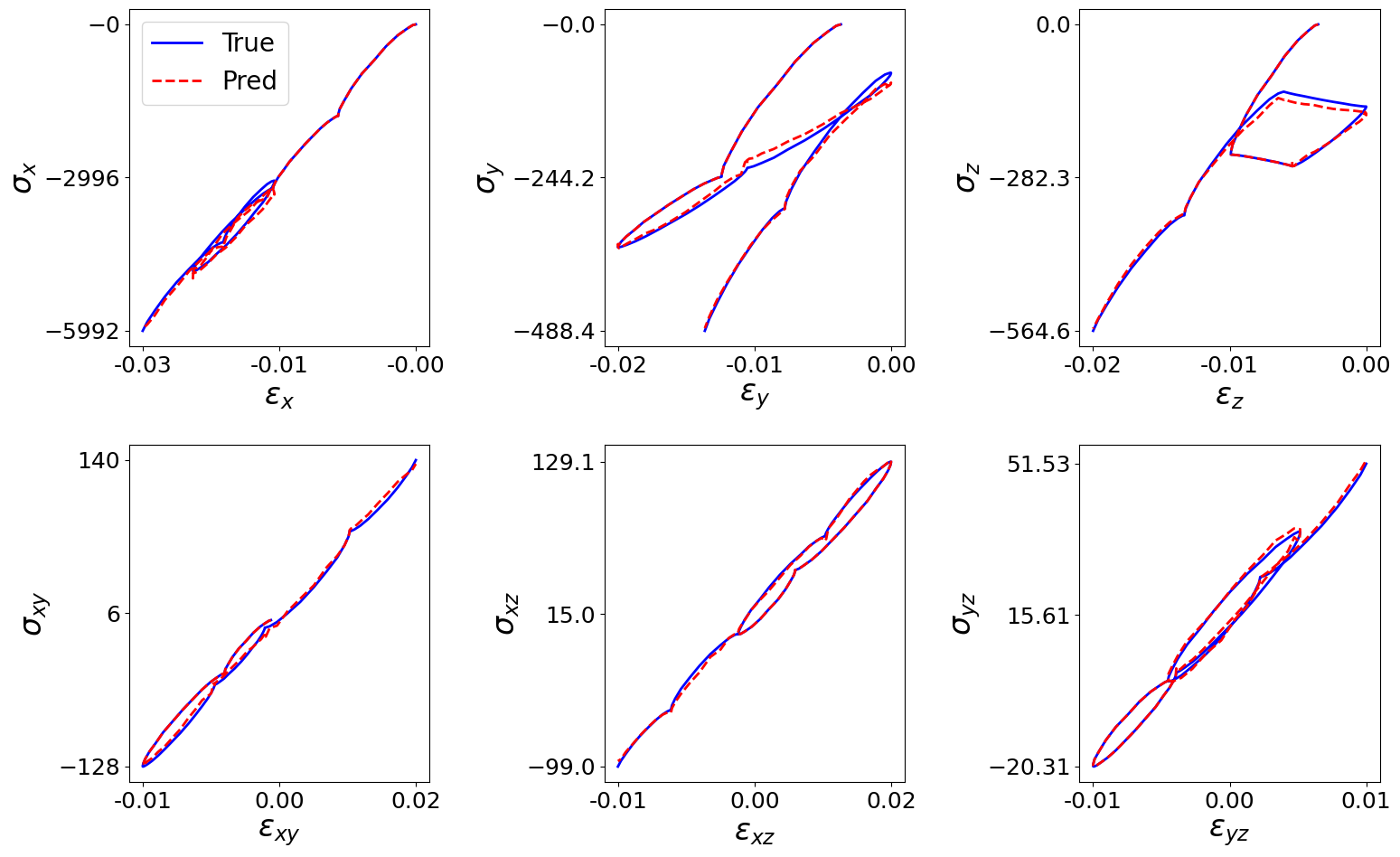}
      \label{fig:case3}
    \end{subfigure}

    \caption{Stress predictions by HANO for three representative test cases associated with the anisotropic
damage model, illustrating its ability to accurately capture complex path-dependent responses.}
    \label{fig:comparison_cases}
  \end{figure}

To validate that the observed improvements of HANO generalize beyond a single loading history, we evaluate three additional test paths randomly selected from the unseen test set~\cite{ge2024data}, as shown in Figure~\ref{fig:comparison_cases}.
Across all cases, HANO consistently demonstrates high fidelity, accurately and robustly capturing both  sharp transition  features and the overall stress evolution. 

\textcolor{blue}{These results underscore the importance of architectural innovations beyond simply  increasing  model complexity, especially for predicting long-term stress evolution in strongly history-dependent materials.
The key contribution of this work is the development of an autoregressive neural operator framework for path-dependent materials, which eliminates the need for latent hidden states used in RNN-based models while delivering substantial performance gains. 
Although this is not an exhaustive architectural study, HANO could be further enhanced through additional refinements.
}

We also remark that while our focus in this Section is on the comparative analysis of different neural operator architectures, all neural operator models tested here outperform the standard neural networks originally used in Ge and Tagarielli~\citep{ge2024data}.

\begin{table}[htbp]
\centering
\begin{tabular}{|l|r|r|r|r|}
\hline
\footnotesize\textbf{Model} & \footnotesize\textbf{Parameter Count} & \begin{tabular}[c]{@{}r@{}}\footnotesize\textbf{Parameter}\\\footnotesize\textbf{Ratio vs. FNO}\end{tabular} & \begin{tabular}[c]{@{}r@{}}\footnotesize\textbf{Performance}\\\footnotesize\textbf{(NRMSE)}\end{tabular} & \begin{tabular}[c]{@{}r@{}}\footnotesize\textbf{Improvement}\\\textbf{vs. FNO}\end{tabular} \\
\hline
FNO & 440,326 & -       & 35.23\% & - \\
\hline
UFNO & 488,634 & +11.0\% & 24.67\% & 30.0\% \\
\hline
HANO & 539,818 & +22.6\% & 21.09\% & 40.1\% \\
\hline
\end{tabular}
\caption{Comparison of baseline FNO, UFNO, and HANO. Parameter ratios and performance improvements are reported relative to the baseline FNO.}
\label{tab:model-comparison}
\end{table}

\begin{table}[h]
\centering
\begin{tabular}{lccc}
\hline
\textbf{Model} & \textbf{Parameter Count} & \textbf{Performance (NRMSE)} \\
\hline
FNO-Large & 545,018 & 33.76\%  \\
UFNO-Large & 548,632 & 24.12\%  \\
HANO & 539,818 & 21.09\%  \\
\hline
\end{tabular}
\caption{\textcolor{blue}{Parameter-matched comparison between FNO, UFNO, and HANO.}}
\label{tab:parameter_matched}
\end{table}

\subsection{Sensitivity to history input length} \label{sec:input_length}

\begin{figure}[htbp]
\centering
\includegraphics[width=0.9\textwidth]{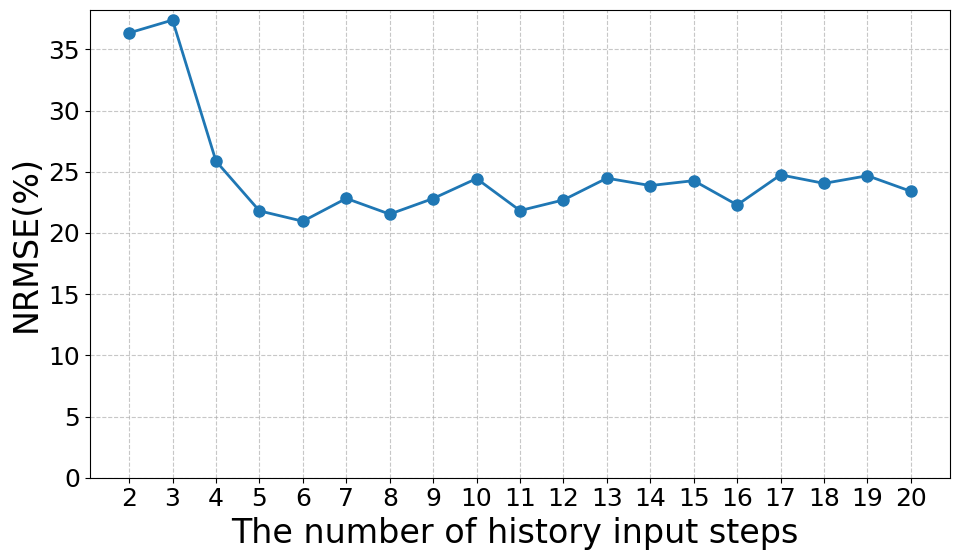}
\caption{Prediction performance of HANO as a function of history input steps. Model accuracy improves with additional steps and stabilizes around six, beyond which further gains are marginal.}
\label{fig:history_steps}
\end{figure}

The number of history input steps ($k$ in Eq. \eqref{eq:reduced-history}) determines how much past information is utilized in stress prediction. 
In this section, we analyze the effect of history input length on model performance for the modified Hashin damage model.
To investigate this, we conduct the training and prediction procedures under different numbers of history input steps, varying from $k = 2$ to $20$. 
Figure \ref{fig:history_steps} shows the NRMSE values as a function of history input length. 
When using a small number of steps ($k = 2$ or $3$), the model exhibits relatively poor performance, with NRMSE around 35 \%.
As the input length increases, prediction accuracy improves rapidly; 
however, the improvement plateaus around $k=6$, 
indicating that additional history information beyond this point yields diminishing returns in prediction accuracy. 
The slight increase and fluctuations in error beyond this point may be attributed to the fixed training protocol, which may require more epochs to accommodate the increased complexity introduced by longer input sequences.

This analysis reveals the importance of selecting an appropriate history length. 
Too few input steps may fail to capture the material’s path-dependent behavior, while an excessive  history steps increase computational cost and training difficulty without significant performance gains. 
For the modified Hashin damage model considered here,
a history length of six strikes a good balance between predictive accuracy and efficiency. 
While the optimal input length is likely constitutive model-dependent, 
necessitating task-specific tuning,
the proposed HANO framework demonstrates robustness across a wide range of input lengths, making it particularly appealing for practical applications.

\textcolor{blue}{An additional consideration worthy of discussion is the potential relationship between the minimum required input length and discretization resolution. For standard machine learning models such as RNNs, finer temporal resolution typically necessitates larger input windows to capture equivalent stress-strain variations, as these models rely on discrete representations where each time step captures less evolutionary information at higher resolutions. While HANO's neural operator framework is designed to learn continuous dynamics independently of discretization (as demonstrated in Sections 4.3.2 and 4.3.3), the theoretical foundation established by Coleman and Gurtin~\cite{coleman1967thermodynamics} suggests that materials with memory fundamentally require a minimum observable window of external variables for accurate constitutive response prediction. This implies that even resolution-invariant models like HANO may encounter limitations when the physical time span covered by the history window becomes insufficient to capture the necessary historical information.}

%% file: Conclusion.tex
\section{Conclusion}\label{sec:conclusion}

In this work, we investigate the use of neural operator (NO) for data-driven modeling of path-dependent nonlinear materials, aiming to  address two fundamental challenges that remain in existing neural network-based approaches: (1) discretization dependence, where model predictions are sensitive to the temporal resolution of the loading path, violating the self-consistency expected of material behavior; and (2) initial state dependence, where  data-driven models require prediction (and hidden variables) to be initialized from a specific reference configuration and struggle to generalize to arbitrary pre-stressed or partially observed loading conditions. 

To address these limitations, we propose the History-Aware Neural Operator (HANO), a novel end-to-end  learning framework that bypasses the need to construct physical or nominal hidden variables,
while achieving robust long-term accuracy in modeling path-dependent constitutive responses.
HANO integrates an attention-enhanced Fourier Neural Operator (FNO) backbone to achieve resolution-invariant 
predictions through spectral learning in continuous function spaces.
It further adopts an autoregressive prediction pipeline that directly conditions on a short window of observable strain–stress history, enabling accurate and consistent forecasts without requiring hidden state initialization.
This design allows HANO to initiate predictions from arbitrary points along the loading path and to handle irregularly sampled loading histories with high fidelity.

The proposed approach is validated through comprehensive numerical studies on both hardening elastoplasticity and a brittle material model with progressive anisotropic damage. 
The results demonstrate that HANO accurately captures complex history-dependent behavior across  diverse material systems,
significantly outperforming baselines such as RNNs and standard NO models.
In the elastoplasticity benchmark, HANO achieves a six-fold accuracy improvement over the RNN counterpart when
tested at a resolution (150 steps per cycle) different from the training resolution (100 steps),  
and a more than ninety-fold improvement  
when forecasting from pre-stressed states without complete loading history. 
In the multidimensional damage model, the proposed HANO model (i.e., HANO$_3$), featuring a U-shaped multiscale architecture, effectively captures both global response trends and localized damage features,
achieving 40.1\% accuracy gain  over the baseline FNO. 
When trained on two-cycle loading data, HANO generalizes well to predict three to five loading cycles with NRMSEs consistently below 0.8\%, 
demonstrating strong extrapolation capability. 
Moreover, HANO maintains high predictive accuracy under varying levels of Gaussian noise in stress inputs, highlighting its robustness for real-world applications involving noisy experimental or field data of varying quality.

This study demonstrates that the proposed HANO framework effectively addresses the two key dependency challenges commonly encountered in data-driven constitutive modeling. 
These findings underscore the importance of accounting for such factors to achieve robust generalization and faithful representation of irreversible material behavior. 
Looking forward, several extensions will be pursued to broaden HANO’s capabilities
as a versatile framework for modeling complex inelastic materials. 
\textcolor{blue}{Owing to its end-to-end learning structure, the HANO-based constitutive model offers potential to be integrated into differentiable numerical solvers to enable gradient-based inverse analysis}, design optimization, and real-time simulation within a unified computational framework~\cite{xue2023jax,he2023hybrid,du2024differentiable}. 
Moreover, 
HANO’s flexible architecture and input–output structure allow it to accommodate a wide range of inelastic material behaviors, including rate-dependent responses (e.g., viscoelastic and viscoplastic materials) and heterogeneous composite systems.  
\textcolor{blue}{While the present study focuses  on purely data-driven modeling, 
HANO also lends itself to incorporating
physics-based constraints and knowledge~\cite{maia2023physically,haghighat2023constitutive,he2022thermodynamically,vlassis2023geometric},
which may further enhance generalizability and predictive accuracy for complex inelastic materials, particularly under sparse or noisy data conditions. }


\textcolor{blue}{Despite its demonstrated advantages, we acknowledge a potential limitation regarding the selection of the history window length $k$, a critical design parameter influenced by material complexity and loading patterns. Our analysis shows that HANO maintains robust performance across a range of $k$ values, with accuracy plateauing beyond a material-specific threshold. This behavior makes the framework practical for diverse applications where complete loading histories may not be available.}

%% file: appendix.tex
\appendix
\section{Stress update algorithm for 1D elastoplasticity}\label{sec:app_A}\label{appx:A}

We consider a classical one-dimensional elastoplastic material model with kinematic hardening, where the total strain $\varepsilon$ is decomposed into elastic strain $\varepsilon^e$ and plastic strain $\varepsilon^p$:
\begin{equation}
    \varepsilon = \varepsilon^e + \varepsilon^p.
\end{equation}
The stress response follows Hooke’s law for elasticity:
\begin{equation}
    \sigma = E \varepsilon^e = E (\varepsilon - \varepsilon^p),
\end{equation}
where $E$ is the Young’s modulus. 
When the material undergoes plastic deformation, the evolution of plastic strain is governed by the Prager kinematic hardening rule, which assumes that the back stress $\alpha$ follows a linear evolution with plastic strain \cite{simo2006computational}:
\begin{equation}
    d\alpha = H d\varepsilon^p,
\end{equation}
where $H$ is the kinematic hardening modulus. The yield function is defined based on the von Mises criterion:
\begin{equation}
    |\sigma - \alpha| - \sigma_y \leq 0.
\end{equation}

The updated stress $\sigma$, back stress $\alpha$, and plastic strain $\varepsilon^{p}$ are obtained based on the classical return-mapping algorithm \citep{simo1985consistent} as described below.

Given a strain increment $\Delta \varepsilon$, the trial elastic predictor is first computed:
\begin{equation}
    \sigma^{\text{trial}} = \sigma_{\text{prev}} + E \Delta \varepsilon.
\end{equation}
If the trial stress satisfies $|\sigma^{\text{trial}} - \alpha_{\text{prev}}| \leq \sigma_y$, the response remains elastic, and the stress is updated as:
\begin{equation}
    \sigma = \sigma^{\text{trial}}, \quad \varepsilon^p = \varepsilon^p_{\text{prev}}.
\end{equation}
Otherwise, if $|\sigma^{\text{trial}} - \alpha_{\text{prev}}| > \sigma_y$, plastic flow occurs, and the plastic strain increment is obtained using the consistency condition:

\begin{equation}
    \Delta \varepsilon^p = \frac{|\sigma^{\text{trial}} - \alpha_{\text{prev}}| - \sigma_y}{E + H} \cdot \text{sign}(\sigma^{\text{trial}}).
\end{equation}
The updated stress and plastic strain are then:
\begin{equation}
    \sigma = \sigma^{\text{trial}} - E \Delta \varepsilon^p, \quad \varepsilon^p = \varepsilon^p_{\text{prev}} + \Delta \varepsilon^p.
\end{equation}
Simultaneously, the back stress is updated following the kinematic hardening rule:
\begin{equation}
    \alpha = \alpha_{\text{prev}} + H \Delta \varepsilon^p.
\end{equation}

\input{appendix_damage}

\section{Material parameters for dataset generation in anisotropic damage modeling}\label{appx:params}
Table~\ref{tab:mech_prop} provides the complete set of mechanical properties used in the phenomenological model for dataset generation.

\begin{table}[htbp]
  \centering
  \begin{tabular}{lll|lll|lll}
    \hline
    \multicolumn{3}{c|}{Elastic Parameters} &
    \multicolumn{3}{c|}{Strength Parameters} &
    \multicolumn{3}{c}{Fracture Energies} \\
    \hline
    $E_{xx}$ (GPa) & 198.2 & &
    $X_T$ (MPa)    & 4222  & &
    $G_{FT}$ (J/mm$^2$) & 12.5 & \\
    $E_{yy}$ (GPa) & 11.2 & &
    $X_C$ (MPa)    & 3800 & &
    $G_{FC}$ (J/mm$^2$) & 12.5 & \\
    $E_{zz}$ (GPa) & 11.2 & &
    $Y_T$ (MPa)    & 122  & &
    $G_{MT}$ (J/mm$^2$) & 1 & \\
    $\nu_{xy}$     & 0.29 & &
    $Y_C$ (MPa)    & 122  & &
    $G_{MC}$ (J/mm$^2$) & 1 & \\
    $\nu_{xz}$     & 0.29 & &
    $S_l$ (MPa)    & 81   & & & & \\
    $\nu_{yz}$     & 0.50 & &
    $S_t$ (MPa)    & 8.7  & & & & \\
    $G_{xy}$ (GPa) & 8.6  & & & & & & & \\
    $G_{xz}$ (GPa) & 8.6  & & & & & & & \\
    $G_{yz}$ (GPa) & 3.7  & & & & & & & \\
    \hline
  \end{tabular}
  \caption{Mechanical properties for the phenomenological model. $E_{xx}, E_{yy}, E_{zz}$ are the elastic moduli; $\nu_{xy}, \nu_{xz}, \nu_{yz}$ are the Poisson’s ratios; $G_{xy}, G_{xz}, G_{yz}$ are the shear moduli; and $G_{FT}, G_{FC}, G_{MT}, G_{MC}$ are the fracture energies associated with the four failure modes.}
  \label{tab:mech_prop}
\end{table}
  
\input{appendix_parameters}

\input{appendix_loss}

%% file: appendix_damage.tex
\section{Anisotropic damage in brittle solids}\label{appx:hashin}

For completeness, we summarize the stress and damage update procedures following the study \cite{ge2024data}, where the associated material parameters used to generate dataset are provided in \ref{appx:params}.

Prior to the initiation of any damage mechanism, the stress–strain relation is
assumed to be orthotropic and elastic, given by
\[
  \sigma_{ij}=C_{ijkl}\,\varepsilon_{kl},
\]
where $\sigma_{ij}$ denotes the Cauchy stress tensor in the
current configuration, $\varepsilon_{kl}$ is the logarithmic strain tensor,
and $C_{ijkl}$ is the fourth-order orthotropic elastic stiffness tensor.
In the material’s local orthotropic basis $(x,y,z)$, the $x$-direction is aligned with the reinforcing fibres (longitudinal axis), whereas the $y$- and $z$-directions span the transverse \(y\text{–}z\) plane that is dominated by the polymer matrix.  
Accordingly, the four modified Hashin criteria~\cite{hashin1980failure} distinguish
\emph{fibre-dominated tension} (FT) and \emph{compression} (FC) along the $x$-direction, and
\emph{matrix-dominated tension} (MT) and \emph{compression} (MC) within the \(y\text{–}z\) plane. Damage initiates when any modified Hashin function $F_I$ ($I \in \mathcal{A} = \{FT,FC,MT,MC\}$) reaches unity,
\begin{align}
F_{FT} &= \left(\frac{\sigma_{xx}}{X_T}\right)^{2}
       + \frac{\sigma_{xy}^{2}+\sigma_{xz}^{2}}{S_l^{2}}
       \ge 1,
\\
F_{FC} &= \left(\frac{\sigma_{xx}}{X_C}\right)^{2}
       \ge 1,
\\
F_{MT} &= \left(\frac{\sigma_{yy}+\sigma_{zz}}{Y_T}\right)^{2}
       + \frac{\sigma_{yz}^{2}-\sigma_{yy}\sigma_{zz}}{S_t^{2}}
       + \frac{\sigma_{xy}^{2}+\sigma_{xz}^{2}}{S_l^{2}}
       \ge 1,
\\
F_{MC} &= \Bigl[\bigl(\tfrac{Y_C}{2S_t}\bigr)^{2}-1\Bigr]
         \frac{\sigma_{yy}+\sigma_{zz}}{Y_C}
       + \frac{(\sigma_{yy}+\sigma_{zz})^{2}}{4\,S_t^{2}}
       + \frac{\sigma_{yz}^{2}-\sigma_{yy}\sigma_{zz}}{S_t^{2}}
       + \frac{\sigma_{xy}^{2}+\sigma_{xz}^{2}}{S_l^{2}}
       \ge 1,
\end{align}
Here $X_T,X_C$ (axial) and $Y_T,Y_C$ (transverse) denote tensile and
compressive strengths; $S_l,S_t$ are longitudinal and transverse shear
strengths. Once $F_I \geq 1$, the associated damage variable $d_I\in[0,1]$ evolves according to
\[
  d_I=\frac{\delta_{I,eq}^{f}(\delta_{I,eq}-\delta_{I,eq}^{0})}{\delta_{I,eq}(\delta_{I,eq}^{f}-\delta_{I,eq}^{0})},\quad
  \delta_{I,eq}^{f}=\frac{2G_I}{\sigma_{I,eq}^{0}},\quad
  \delta_{I,eq}^{0}=\frac{\delta_{I,eq}}{\sqrt{F_I}},\quad
  G_I=\tfrac12\sigma_{I,eq}^{f}\varepsilon_{I,eq}^{f}l_c,
\]
with $l_c$ the characteristic element length (cubic root of element volume).  
Fibre and matrix degradation combine as
\[
  d_f=(1-d_{FT})(1-d_{FC}),\qquad d_m=(1-d_{MT})(1-d_{MC}),
\]

Given total strain $\varepsilon_{kl}^{n+1}$ and damage state $d_I^{\,n}$, the undamaged trial stress is computed:
\[
 \sigma_{ij}^{\text{trial}}=C_{ijkl}\bigl(\varepsilon_{kl}^{n+1}-\varepsilon_{kl}^{p,n}\bigr).
\]
The trial stress is used to evaluate the damage initiation functions $F_I$ and identify active modes:
\[
\bar{\mathcal{A}}=\{\,I \mid F_I^{\text{trial}} \geq 1, I \in \mathcal{A}\}.
\]
For each active mode $I\in\bar{\mathcal{A}}$, the equivalent displacement is determined:
\[
\delta_{I,eq}^{n+1}=\sqrt{S_I^{\alpha\beta}\sigma_{\alpha}^{\text{trial}}\sigma_{\beta}^{\text{trial}}},
\]
and the damage variable is updated according to:
\[
d_I^{\,n+1}=\frac{\delta_{I,eq}^{f}\bigl(\delta_{I,eq}^{n+1}-\delta_{I,eq}^{0}\bigr)}
                {\delta_{I,eq}^{n+1}\bigl(\delta_{I,eq}^{f}-\delta_{I,eq}^{0}\bigr)},
\]
with $\delta_{I,eq}^{0}=\delta_{I,eq}^{n+1}/\sqrt{F_I^{\text{trial}}}$. For inactive modes, $d_I^{\,n+1}=d_I^{\,n}$.
The combined degradation factors are computed as:
\[
d_f=(1-d_{FT}^{\,n+1})(1-d_{FC}^{\,n+1}), \quad
d_m=(1-d_{MT}^{\,n+1})(1-d_{MC}^{\,n+1}),
\]
and the stress is updated using the degraded stiffness:
\[
\sigma_{ij}^{\,n+1}=\bigl[(1-d_f)C^{f}_{ijkl}+(1-d_m)C^{m}_{ijkl}\bigr]
                    \bigl(\varepsilon_{kl}^{n+1}-\varepsilon_{kl}^{p,n}\bigr),
\]
where $C^{f}_{ijkl}$ and $C^{m}_{ijkl}$ are stiffness partitions attributed to fibre and matrix, respectively.
The consistent tangent operator for the next iteration is:
\[
 \mathbb{C}_{ijmn}^{\text{alg}}
 =C_{ijkl}^{d}\delta_{km}\delta_{ln}
 -\sum_{I\in\mathcal{A}}
      \frac{\partial d_I}{\partial \delta_{I,eq}}
      \frac{\partial \delta_{I,eq}}{\partial \varepsilon_{mn}}\,
      C^{I}_{ijkl}\sigma_{kl}^{n+1},
\]
with degraded stiffness $C^{d}_{ijkl}=(1-d_f)C^{f}_{ijkl}+(1-d_m)C^{m}_{ijkl}$ and:
\[
\frac{\partial \delta_{I,eq}}{\partial \varepsilon_{mn}}
  =\frac{S_I^{\alpha\beta}C_{\alpha mn}\sigma_{\beta}^{\,n+1}}
         {\delta_{I,eq}^{\,n+1}}.
\]
The non-zero $S_I^{\alpha\beta}$ components needed for the projector are:
\[
\renewcommand{\arraystretch}{1.05}
\begin{array}{c|cccccc}
I & S_{xx} & S_{yy} & S_{zz} & S_{xy} & S_{xz} & S_{yz} \\ \hline
FT & 1/X_T^{2} & 0 & 0 & 1/S_l^{2} & 1/S_l^{2} & 0 \\
FC & 1/X_C^{2} & 0 & 0 & 0 & 0 & 0 \\
MT & 0 & 1/Y_T^{2} & 1/Y_T^{2} & 1/S_l^{2} & 1/S_l^{2} & -1/S_t^{2} \\
MC & 0 & 1/(4S_t^{2}) & 1/(4S_t^{2}) & 1/S_l^{2} & 1/S_l^{2} & 1/S_t^{2}
\end{array}
\]

%% file: appendix_parameters.tex
\section{\textcolor{blue}{Parameter-Matched Comparison with RNN Models}}
\label{appx:parameter_matched_rnn}

\textcolor{blue}{To ensure a fair comparison between HANO and RNN-based models, we conducted additional experiments with parameter-matched configurations. Following the experimental setup described in Section~\ref{sec:start_data}, we used the same 1D elastoplastic material model with identical training and testing datasets to evaluate model performance under controlled conditions.}

\textcolor{blue}{The RNN models are configured as follows: RNN\(_1\) (77,121 parameters) uses a 1D strain input through Input Projection (1→64) + 3×GRU(64→64) + Output Projection (64→32→1), while RNN\(_2\) (77,185 parameters) employs the same architecture but modifies the input projection to Linear(2→64) to accommodate 2D input (strain + stress), differing by only 64 parameters. For the HANO variants, the original configuration uses width=64, spectral modes=5, and 6 layers (3 Fourier + 3 AEUF), totaling 455,685 parameters. To enable parameter-matched comparison, we created HANO\(_{\text{reduced}}\) with width=32, spectral modes=5, and 4 layers (L=2 Fourier + M=2 AEUF), totaling 77089 parameters to match the RNN models.}

\textcolor{blue}{All models were trained using identical protocols, including progressive scheduled sampling and noise injection strategies as described in Section~3.3.3. The evaluation was conducted on the same two test sets: Testset I (full loading history with undeformed start) and Testset II (partial loading history with deformed start). Table~\ref{tab:parameter_matched_rnn_comparison} presents the NRMSE results for all four models.}

\begin{table}[ht!]
\centering
\begin{tabular}{lcccc}
\hline
\textbf{Loading Condition} & \textbf{RNN\(_1\)} \cite{gorji2020potential} & \textbf{RNN\(_2\)} & \textbf{HANO} & \textbf{HANO\(_{\text{reduced}}\)} \\ 
\hline
\textbf{Parameters} & 77,121 & 77,185 & 455,685 & 77,089 \\
\hline
Testset I & 0.030 & 0.023 & 0.006 & 0.009 \\
Testset II & 0.359 & 0.106 & 0.004 & 0.008 \\ 
\hline
\end{tabular}
\caption{\textcolor{blue}{Parameter-matched comparison of the NRMSE error between RNN variants and HANO models for different test sets.}}
\label{tab:parameter_matched_rnn_comparison}
\end{table}

\textcolor{blue}{The results demonstrate that even when HANO's parameter count is reduced to match the RNN models (~77k parameters), HANO\(_{\text{reduced}}\) achieves substantially better performance than both RNN variants. Specifically, on Testset I, HANO\(_{\text{reduced}}\) achieves an NRMSE of 0.009 compared to 0.023 for RNN\(_2\), representing a 60\% improvement. On the more challenging Testset II, HANO\(_{\text{reduced}}\) achieves an NRMSE of 0.008 compared to 0.106 for RNN\(_2\), representing a 92\% improvement. These results confirm that HANO's superior performance stems from its architectural innovations—particularly the neural operator framework and attention-enhanced U-Fourier layers—rather than simply increased model capacity.}

%% file: appendix_loss.tex
\section{\textcolor{blue}{Training and Validation Losses}}\label{appx: appendix_loss_curve}
\begin{figure}[h!]
\centering
\includegraphics[width=0.8\textwidth]{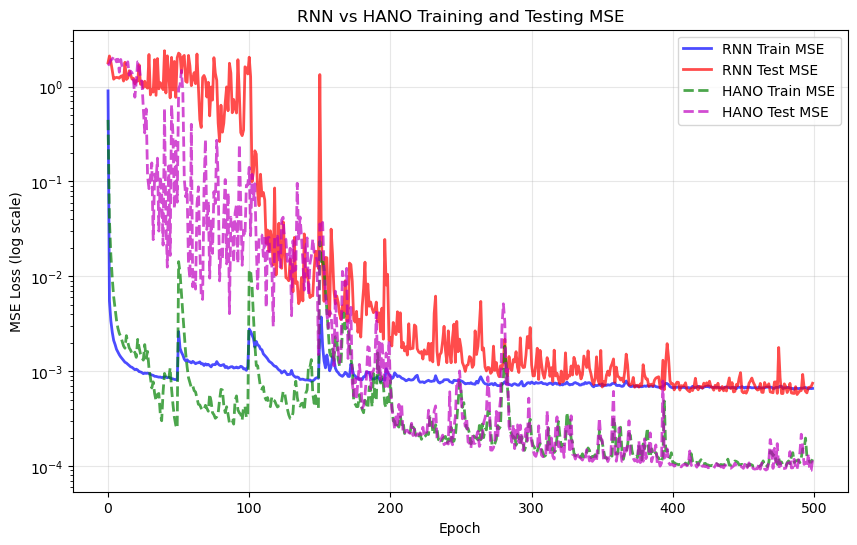}
\caption{\textcolor{blue}{Training and validation loss curves for the HANO and RNN\(_2\) model on the variable-resolution elastoplastic material dataset described in Section 4.3.3.}}
\label{fig:hano_rnn_training_curves}
\end{figure}

\textcolor{blue}{The training and validation losses corresponding to the experiments in Section \ref{sec:resolution_training} are provided in this Appendix. Figures \ref{fig:hano_rnn_training_curves} shows the evolution of the Mean Squared Error (MSE) loss during training for the HANO model and the RNN\(_2\) baseline, respectively. 
The HANO model exhibits stable and efficient training behavior, with no signs of overfitting or underfitting.
Both training and validation losses converge to similar final values around $10^{-4}$, indicating that the model has sufficient capacity to learn the underlying patterns while generalizing well to unseen data.
In the initial training phase (approximately the first 200 epochs), there is a noticeable gap between training and validation losses, which is primarily attributed to the teacher forcing mechanism used in the progressive training strategy. As the scheduled sampling gradually transitions from teacher forcing to self-reliant prediction, the training and validation curves align more closely, demonstrating the effectiveness of this training approach.}

\textcolor{blue}{The HANO model shows smooth and stable convergence throughout the training process. The training loss decreases consistently from approximately $10^{-1}$ to $10^{-4}$ over 500 epochs, while the validation loss follows a similar trajectory with minimal fluctuation. The model reaches stable convergence around epoch 400, with both losses plateauing at similar low values, confirming robust learning without overfitting.}

\textcolor{blue}{In contrast, the RNN\(_2\) baseline exhibits different training dynamics. While the training loss converges smoothly to approximately $10^{-3}$, there is a persistent gap between training and validation losses throughout the training process. The validation loss shows more fluctuation compared to HANO, particularly in the later stages of training, suggesting less stable generalization behavior. The final validation loss remains approximately one order of magnitude higher than HANO's final loss, indicating inferior performance.}

